\documentclass[10pt,twocolumn,letterpaper]{article}

\usepackage[pagenumbers]{wacv} %

\usepackage{comment}

\newif\ifshowTODOs
\showTODOsfalse %

\newcommand{\TODO}[1]{\ifshowTODOs\textbf{\color{red}[TODO: #1]}\fi}
\newenvironment{TODOB}
    {%
        \ifshowTODOs
            \textbf{\color{red}[TODOBEGIN]}\par\medskip\bgroup\normalcolor
        \else
            \setbox0=\vbox\bgroup
        \fi
    }
    {%
        \ifshowTODOs
            \egroup\par\medskip\textbf{\color{red}[TODOBEND]}%
        \else
            \egroup\setbox0=\box0
        \fi
    }

\definecolor{wacvblue}{rgb}{0.21,0.49,0.74}
\usepackage[pagebackref,breaklinks,colorlinks,allcolors=wacvblue]{hyperref}

\def\wacvPaperID{1667} %
\def\confName{WACV}
\def\confYear{2026}

\usepackage{float}
\usepackage{caption}
\usepackage{lipsum}
\usepackage{booktabs}
\usepackage{siunitx}
\usepackage{caption}
\usepackage{multirow}
\usepackage{booktabs}
\usepackage{dsfont}
\PassOptionsToPackage{table}{xcolor}
\usepackage{xcolor}
\usepackage{colortbl}
\usepackage{comment}
\title{HANDI: Hand-Centric Text-and-Image Conditioned Video Generation}

\author{
Yayuan Li\textsuperscript{1,}\thanks{Equal contribution} \quad Zhi Cao\textsuperscript{1,}\footnotemark[1] \quad Jason J. Corso\textsuperscript{1,2} \\
\textsuperscript{1}University of Michigan \quad \textsuperscript{2}Voxel51 \\
\href{https://excitedbutter.github.io/project_page}{Project Page} \\
}

\begin{document}

\twocolumn[{%
\renewcommand\twocolumn[1][]{#1}%
\maketitle
\begin{center}
    \centering
    \captionsetup{type=figure}
    \includegraphics[width=.95\textwidth
    ]{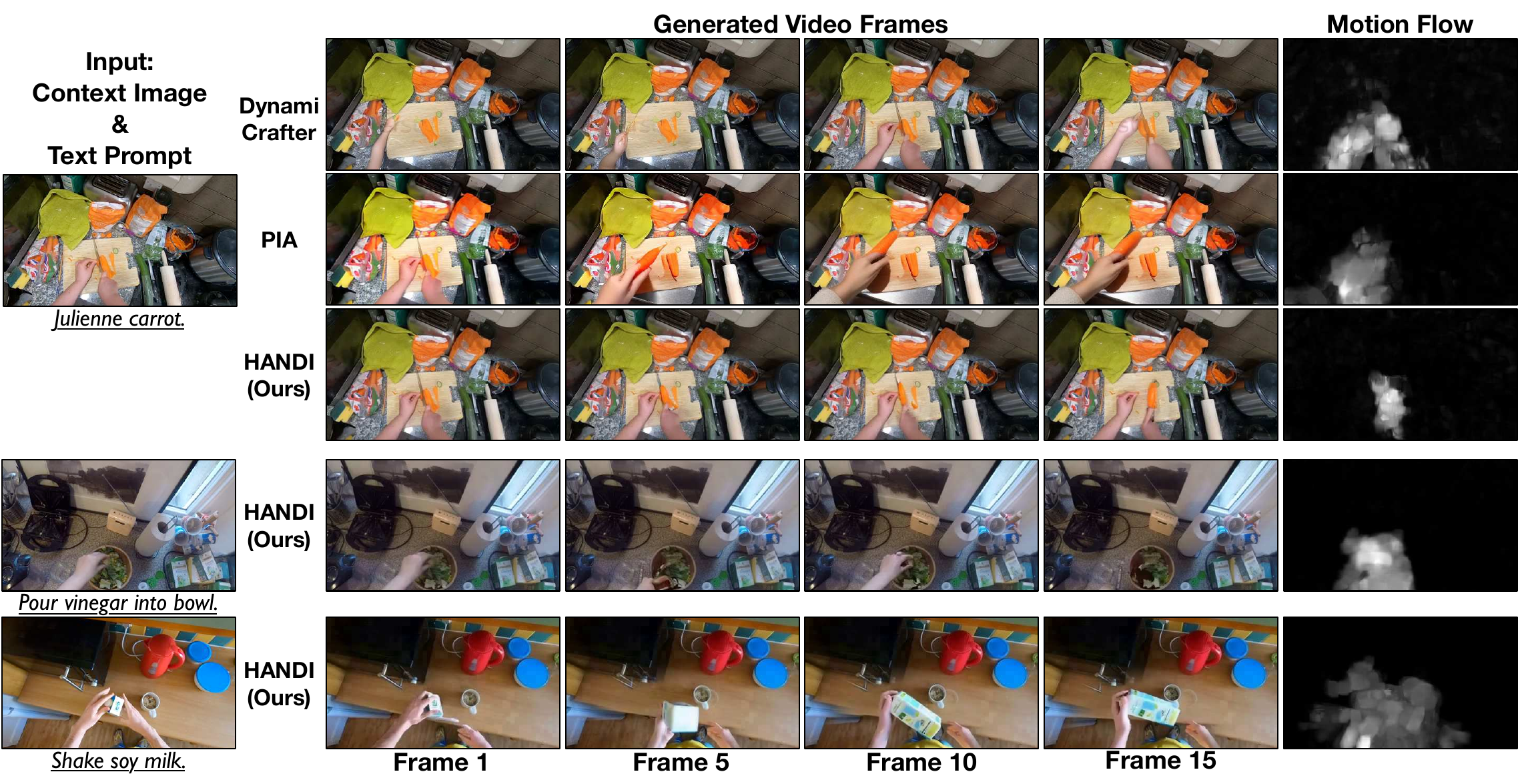}
    \captionof{figure}{
Illustration of our proposed Hand-Centric Text-and-Image Conditioned Video Generation (HCVG). Given an image for context and an action text prompt, our method generates video frames with precisely refined hand appearance and motion, overcoming challenges like unreasonable motion in backgrounds. Unlike baselines that extend unnecessary motion to background with rough hand structure, our approach produces motion in more reasonable area, as shown in the Motion flow visualization and refined hand details.
    }
    \label{fig:intro}
\end{center}%
}]

\begin{abstract}
Despite the recent strides in video generation, state-of-the-art methods still struggle with elements of visual detail.  One particularly challenging case is the class of videos in which the intricate motion of the hand coupled with a mostly stable and otherwise distracting environment is necessary to convey the execution of some complex action and its effects.  To address these challenges, we introduce a new method for video generation that focuses on hand-centric actions.  Our diffusion-based method incorporates two distinct innovations.  First, we propose an automatic method to generate the motion area---the region in the video in which the detailed activities occur---guided by both the visual context and the action text prompt, rather than assuming this region can be provided manually as is now commonplace.  Second, we introduce a critical Hand Refinement Loss to guide the diffusion model to focus on smooth and consistent hand poses.  We evaluate our method on challenging augmented datasets based on EpicKitchens and Ego4D, demonstrating significant improvements over state-of-the-art methods in terms of action clarity, especially of the hand motion in the target region, across diverse environments and actions.

\vspace{-1em}
\end{abstract}

\section{Introduction}

Instructional media have become the de-facto manual for everything from “how to poach an egg” videos to full laparoscopic surgery walkthroughs~\cite{wang2016multi,YoAyCaSI2022}.  
More than half of YouTube users now search specifically for how-to content~\cite{smith2018many}.  
Within this vast stream, the clips most useful for both humans and robots are those that capture \emph{goal-oriented hand manipulation}: fine, dexterous interactions between hands, tools, and objects performed in ordinary, cluttered environments~\cite{zhou2018towards,damen2022rescaling,grauman2022ego4d,wang2023holoassist}.  
Such footage accelerates human skill acquisition and has proved invaluable for imitation-based policy learning~\cite{yang2023learning,jain2024vid2robot,chanesane2023vip}.

Obtaining task-specific videos on demand remains difficult.  
Retrieval systems often return clips recorded in different kitchens, workshops, or lighting, breaking visual continuity with the user’s scene~\cite{ashutosh2024detours,grauman2024ego}.  
Single-frame generators can preserve context but sacrifice temporal dynamics~\cite{lai2023lego,souvcek2024genhowto}.  
Recent text-and-image-to-video (TI2V) models attempt to offer both context and motion~\cite{ni2024ti2v,zhang2024pia}, yet they are designed for web benchmarks featuring centred subjects against clean backgrounds~\cite{bakr2023hrs,basu2023editval,huang2023t2i,lee2024holistic,saharia2022photorealistic,wang2023imagen,aifanti2010mug,schuhmann2021laion,bain2021frozen,chen2024panda,xue2022advancing}; when confronted with real household footage they hallucinate background motion and render low-fidelity hands, especially when the hands occupy only a small portion of the frame (see~\cref{fig:intro}).  
Bridging this gap is essential for practical human-assistive and robot-assistive scenarios.

Motivated by these limitations, we cast the task as \emph{hand-centric video generation (HCVG)}.  
Given (i) a single RGB scene image that shows the workspace \emph{before} execution and (ii) a text command describing the desired manipulation, HCVG requires synthesising a video that  
(a) alters pixels only inside the \emph{motion area}—the small, task-specific region encompassing the hand, tool, and target object—while leaving the surrounding context untouched;  
(b) produces anatomically correct, temporally smooth hand motion; and  
(c) realises any downstream object-state change required by the action (e.g., transforming a whole carrot into diced pieces after \textit{cut carrot}; see ~\cref{fig:intro}).  
In another word, HCVG emphasize the same input and output modality as TI2V but focuses on animating high-quality hands from challenging, unedited real-world images.
Note that, unlike instruction-based TI2V, which enjoys more information (e.g., the motion area or action trajectory) from the user~\cite{brooks2023pix2pix,dai2023fine,yin2023dragnuwa}, HCVG requires less user burden and requires the generator itself to localise and track this region in cluttered scenes where the hand often occupies less than a quarter of the frame.

To meet these demands we introduce \underline{HAN}d-centric \underline{DI}ffusion (\textsc{HANDI}), a two-stage 3-D diffusion framework.  
Stage 1 automatically predicts a spatio-temporal motion-area (MA) mask from the image–text pair, using a lightweight video-diffusion backbone
.  
We develop a pipeline to automatically prepare the pseudo MA masks from training videos for training. 
Stage 2 conditions the backbone~\cite{ho2022video,rombach2022high} on the predicted mask to generate the instructional video clip. This stage optimises a \emph{Hand-Refinement Loss} for hand-pose consistency. HRL supervises by comparing the generated and groundtruth hand skeletons, across all generated frames, introducing explicit hand structure modeling among plain pixels. 
This two-stage MA-aware generation framework and HRL allow \textsc{HANDI} to render high-fidelity hands even in the presence of severe clutter.

To benchmark HCVG, we use two large-scale hand action datasets, Ego4D~\cite{grauman2022ego4d} and EpicKitchens~\cite{damen2022rescaling}, featuring various unedited real-world visual backgrounds. In addition to various standard video generation metrics, we propose two additional metrics focusing on good MA and hand appearance. We first compare with SOTA video generation methods to show HANDI's effectiveness, demonstrating our method's strong performance for HCVG.
Then, we take a closer look at model performance along key axes: (1) the performance of samples with large MA and motion complexity defined by hand movement, (2) the ability to capture the fine-grained motion of the hands, (3) the performance within the MA versus the whole frame, and (4) compute time needed given the two-stage nature of our method. We show that our model outperforms SOTA baselines while maintaining computation efficiency. 

Our key contributions are two-fold:

\begin{enumerate}
\item We identify the unique aspects of hand-centric video generation (HCVG) and propose a new method, HANDI, that is able to address these unique challenges.
\item We demonstrate that HANDI is able to outperform state of the art video generation from image and text-prompt methods, along all measurement axes
\end{enumerate}

\section{Related Work}
\noindent\textbf{Position of our work.}  
\textsc{HANDI} unifies the desirable properties of several research threads while removing their respective limitations.  
Unlike \emph{image} generation work~\cite{lai2023lego,souvcek2024genhowto}, our method produces dynamics---temporally coherent demonstrations.  
Unlike text-only video models (I2V), which ignore the user’s scene and therefore hallucinate context~\cite{zhang2024show1,wang2024lavie}, \textsc{HANDI} conditions on the real workspace image and keeps the visual context like object appearance matched.  
In contrast to prompt-style Text-Image-to-Video (IT2V) generation work that require masks or trajectories from the user~\cite{dai2023fine,yin2023dragnuwa}, \textsc{HANDI} does not require those scribbles and lowers user burden, enabling scalable deployment in cluttered, real-world environments where the hand occupies only a tiny fraction of the view.
Finally, compared with standard TI2V systems that animate the whole frame and blur small manipulators~\cite{ni2024ti2v,zhang2024pia}, we explicitly localise the motion area and refine 3-D hand pose, yielding the right amount of high-fidelity articulation.  

\textbf{Instructional-video corpora.}  
Large egocentric datasets such as Assembly101~\cite{sener2022assembly101}, YouCook2~\cite{zhou2018towards}, Meccano~\cite{ragusa2021meccano}, HoloAssist~\cite{wangholoassist}, Ego4D~\cite{grauman2022ego4d} and EPIC-Kitchens~\cite{damen2022rescaling} capture sequences of human-object interaction and drive progress in action recognition and imitation learning.  Their footage is \emph{unedited}: hands are small and backgrounds cluttered.  

\noindent\textbf{Image- and video-based visual instruction.}  
Context-preservation is the key to providing instruction. Viewpoint and appearance mismatches increase cognitive load for humans~\cite{chandler1991cognitive} and degrade robot policies~\cite{mandikal2021dexvip}. 
Generating static, context-aware instructional images has been explored by LEGO~\cite{lai2023lego} and GenHowTo~\cite{souvcek2024genhowto}, but images lack temporal cues essential for manipulation.  Video generation conditioned on text alone has advanced rapidly~\cite{zhang2024show1,wang2024lavie}, yet these models produce videos with mismatched context.
Text-and-image-to-video (TI2V) systems such as TI2V-Zero~\cite{ni2024ti2v} and PIA~\cite{zhang2024pia} add scene context but still treat the whole frame as a canvas, resulting in background motion and distorted hands.  Recent instruction-based editors (e.g., Pix2Video~\cite{ceylan2023pix2video}, AnimateAnything~\cite{dai2023fine}, DragNUWA~\cite{yin2023dragnuwa}) mitigate distractions by \emph{requiring} user-supplied masks or keypoint trajectories, limiting scalability.

\noindent\textbf{Motion-area localisation and hand fidelity.}  
When the manipulation locus is tiny—often \(<\!25\%\) of the frame—automatic localisation becomes critical.  Prior work either relies on large-scale domain randomisation \cite{du2024learning,soni2024videoagent} or manual markup~\cite{dai2023fine,yin2023dragnuwa} and still exhibits camera drift.  General-purpose video diffusers such as VideoCrafter~\cite{chen2024videocrafter2} or latent transformers~\cite{nichol2021improved} improve realism but do not model articulated hand structure, leading to flicker and anatomical errors.  Hand-specific refiners for \emph{images} (HandDiffuser~\cite{narasimhaswamy2024handiffuser}, HandRefiner~\cite{lu2024handrefiner}) do not address temporal coherence.

\section{Hand-Centric Video Generation}
\label{sec:method}

\noindent\textbf{Problem Statement}\quad
Formally, we define the hand-centric text-and-image conditioned video generation (HCVG) problem as follows. Given an RGB image $I \in \mathds{R}^{H \times W \times 3}$ containing a manipulator
(a hand or hands)
ready to perform an action, and a text prompt $T$ to describe the action, an HCVG method will output an RGB video clip $V \in \mathds{R}^{L \times H \times W \times 3}$ demonstrating how the action in $T$ is done in the same visual environment shown in $I$.   We reserve notation $I$ for specifying this \textit{visual context}.   In the remainder of the paper, we adopt a shorthand notation when specifying the dimensions of matrices: let $\boxtimes \doteq H \times W$ and it subsumes neighboring $\times$.  So, $I\in\mathds{R}^{\boxtimes\,3}$ and $V\in\mathds{R}^{L\,\boxtimes\,3}$.

\noindent\textbf{Training Data Expectations}\quad
Given the rich instructional and action-oriented video corpora~ \cite{sener2022assembly101,zhou2018towards,ragusa2021meccano,wangholoassist,miech2019howto100m,marin2018recipe1m+,damen2022rescaling,grauman2022ego4d}, we can assume a corpus of training videos for HCVG.  Each training sample will be comprised of a training video $V \in \mathds{R}^{(L+1)\,\boxtimes\,3}$ and a text prompt $T$ describing the goal-oriented activity in the video.  The first frame of $V$ serves as the visual context $I$ for HCVG.  The remaining $L$ frames are the ideal video generation output given $I$ and $T$.

\noindent\textbf{Method Overview}\quad

\TODO{A workflow figure}
We adapt SOTA video latent diffusion techniques~\cite{ni2023conditional, esser2023structure, wang2024videocomposer} to be the core of our solution to the HCVG problem.  Our model has two stages: stage 1 generates the motion area mask and stage 2 generates the video.  However, we use the same diffusion backbone for both stages, simplifying our discussion. 

\noindent\textit{Action Text Embedding}\quad Let us first explain how we represent that action text prompt $T$ to allow us to effectively condition the generation process on it.  To generate the \textit{action text embedding}, $E_{text}$, we first expand $T$ using a pre-trained instruction tuning module~\cite{lai2023lego} to generate a rich textual description based on the evidence that this improves conditional diffusion models~\cite{lai2023lego, souvcek2024genhowto, zhang2022motiondiffuse, cen2024text_scene_motion}.  After expansion, we use the CLIP text embedding~\cite{radford2021learning} to compute $E_{text} \in \mathds{R}^{N \times d}$ where $N$ is the number of tokens and $d$ is the feature dimension.

\noindent\textit{Training}\quad Consider a training video $V \in \mathds{R}^{(L+1)\,\boxtimes\,3}$ and its associated action text embedding $E_{text}$.
We process $V$ frame-by-frame with the pre-trained VAE encoder from Stable Diffusion v1.5~\cite{rombach2022high} to compute the latent video, which is common practice~\cite{dai2023fine, esser2023structure, ho2022video, singer2022make, wang2024videocomposer}. 
After concatenating the prior or generated motion area mask (described in \S\ref{sec:rom}), we arrive at $z \in \mathds{R}^{(L+1)\,\boxtimes\,(c+1)}$, where $c$ is the latent space dimensionality. 
We remove the extra one timestamp to produce the $L$-frame target video.

For each diffusion training forward step, following standard practices~\cite{rombach2022high, lu2022dpm, lu2022dpmpp}, we prepare the noisy latent video $z_{t}$ for a randomly sampled $t \in {1, ..., \tau}$ where $\tau$ is a predefined max step, according to standard noise-adding process~\cite{ho2020denoising, nichol2021improved}:
\begin{align}
z_t = \sqrt{\prod_{i=1}^t (1 - \beta_i)} \, z_0 + \sqrt{1 - \prod_{i=1}^t (1 - \beta_i)} \, \epsilon , 
\label{eq:adding_noise}
\end{align}
where $\beta_i$ is a predefined coefficient that controls noise intensity at step $i$ and $\epsilon \sim \mathcal{N}(0, \mathbf{I})$
is the sampled Gaussian noise.
Then, we use a 3D UNet-based Noise Predictor~\cite{dai2023fine}, aligned with recent video generation work~\cite{singer2022make, esser2023structure, wang2024videocomposer, ho2022video, rombach2022high}, to predict the noise $\mathbf{\hat{\epsilon_{t}}}$ during the denoising process. 
The noise predictor takes the current noisy latent sample $z_{t}$ and conditions on the $E_{text}$ for prediction. Each of its blocks contains convolution layers~\cite{lecun1998gradient} for spatial information extraction at the per frame level, temporal convolution layers~\cite{lea2017temporal, varol2017long} and temporal attention layers~\cite{yan2019stat} for information fusion across all timestamps at video level, and extra cross attention layers~\cite{vaswani2017attention} to fuse information from $E_{text}$.

\noindent\textit{Inference}\quad
Starting with visual context $I$, we produce $z_{context}$ from the pre-trained VAE encoder (also by Stable Diffusion v1.5~\cite{rombach2022high}). Following~\cite{dai2023fine}, we duplicate $z_{context}$ for $L$ times and concatenate the prior or generated MA mask, depending on stage (\S\ref{sec:rom}). Then we apply~\cref{eq:adding_noise} at the max noise step $\tau$ to produce the initial noisy latent video $z'_{\tau}$ as the beginning of denoising process. The denoising process during inference iterates $\tau$ steps. At each denoising step $t$ the trained stage-specific noise predictor model conditionally predicts the noise $\hat{\epsilon}_{t}$ from $z'_{t}$, the same as training process. We apply the DPM++ solver~\cite{lu2022dpmpp,lu2022dpm} to calculate the noisy latent video at previous timestamp $z'_{t-1}$ iteratively for $\tau$ times. 
After $\tau$ steps of denoising, a clean latent video $z'_{0}$ is produced. The pretrained VAE decoder~\cite{rombach2022high} is applied per-frame to produce the generated motion area mask video or generated video in pixel space.  We follow Dai et al.~\cite{dai2023fine} to explicitly enforce the mask after the generation process, replacing the background pixels from the context image $I$.

\vspace{1ex}
\noindent\textbf{Innovations}\quad
Extending this structure, we introduce two main innovations. 
First (\S\ref{sec:rom}), we develop a spatially cued approach that automatically learns the motion area mask given visual context $I$ and text prompt $T$.  This is critical for HCVG as the video generation needs to focus in on the pertinent area of the video while leaving the rest of it undisturbed to avoid distraction.  Furthermore, it encourages the implicit learning of object state changes that frequently occur in our class of videos.
Second (\S\ref{sec:handloss}), we propose a novel Hand Refinement Loss tailored to HCVG. 
This loss greatly helps the generated video deliver higher quality critical details of action information since the hand is the key element when demonstrating actions in this scenario. We describe these two innovations in more detail next, followed by their integration into the full loss in Sec. \ref{sec:fullloss}.

\subsection{Automatically Estimating the Motion Area}
\label{sec:rom}

Hand-centric video generation must focus its effort on the spatiotemporal region that will contain the actual hand-object interaction, to avoid manipulating the visual context and adding distraction in the background.  Hence, a spatiotemporal mask is needed.
Although past work has supported the use of a human-supplied spatial mask
in video generation~\cite{dai2023fine}, 
we focus on automatically generating the mask because it is less onerous and more flexible.
The first stage of our conditional diffusion model must produce a mask video that includes any and all parts of the video that may be used in the execution of the actor given visual context $I$ and action text prompt $T$, which are used during the diffusion process.

\begin{figure}[t]
\centering
\includegraphics[width=0.95\linewidth]{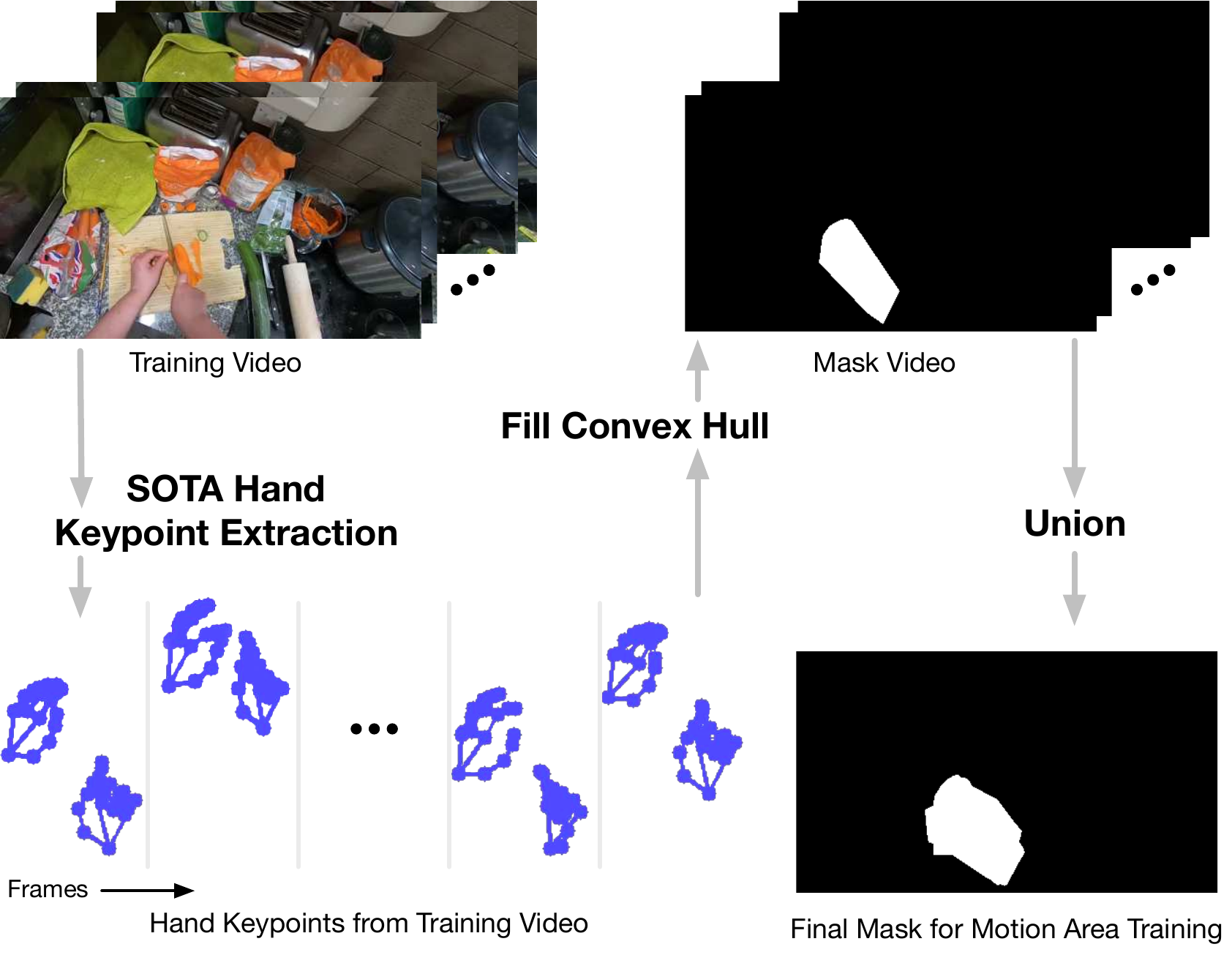}
\vspace{-2ex}
\caption{Training motion area masks are automatically created from the training videos.  We flood fill the convex hulls of the hand regions, and then take the set-union of these over all frames.
}
\label{fig:mask}
\vspace{-2ex}
\end{figure}

For the MA generation, we assume that the relevant regions of activity map to the extent of the hands in the training videos. 
We hence automatically annotate the motion area in a training video using off-the-shelf SOTA hand keypoint detection~\cite{lugaresi2019mediapipe}, as shown in~\cref{fig:mask}.
 
Concretely, given a training video $V \in \mathds{R}^{(L+1)\,\boxtimes\,3}$ (including the context image as the first frame), we apply the SOTA hand keypoint detector, denoted as $\Upsilon$, that outputs, for each frame $l = [0, 1, 2, ... L]$, a set of hand keypoints $P_{l} \in \mathds{R}^{J_{l} \times 2}$ where $J_{l}$ is the number of 2D key points detected for frame~$l$. For each $P_{l}$, we floodfill their convex hull to delimit a region as the mask $M_{l}$ at frame $l$. We then apply union over masks for all frames to produce the final MA mask $M_{video} \in [0,1]^\boxtimes$ for training.  

During training of the stage 1 noise predictors, these mask videos are used as the targets.   Since we use the same diffusion model structure for both stages, we also supply a prior mask during stage 1 training.  This prior mask is computed as the normalized coverage over all groundtruth masks in the training set, indicating the prior distribution of the MA.
During stage 2, the soft region masks outputted from stage 1, post-processed with morphology, are incorporated as part of the conditioning for the video generation.  
Our experiments show the high impact of the MA (\S\cref{sec:exp}).

\subsection{Hand Refinement Loss}
\label{sec:handloss}

\begin{figure}[t]
\centering
\includegraphics[width=0.85\linewidth]{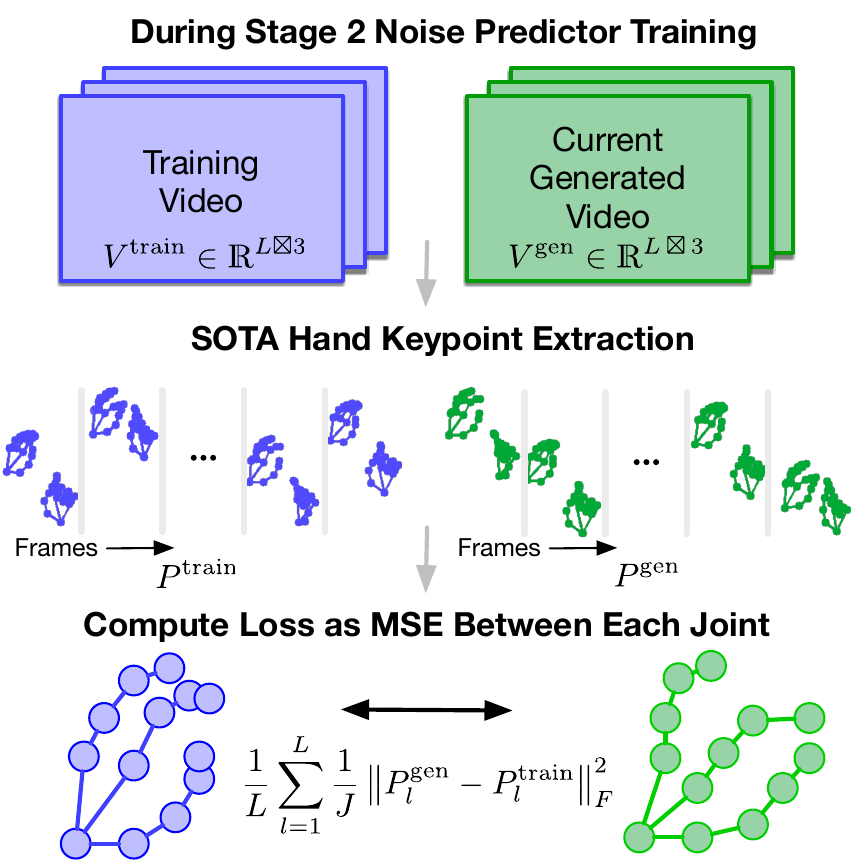}
\vspace{-1.5ex}
\caption{Illustrates the Hand Refinement Loss that drives the stage 2 noise predictor to focus on the fine-grained detailed of the hand pose in the context of the interaction. Bottom pose is for illustration only.  Our representation has 21 joints per hand.}
\label{fig:handpose}
\vspace{-2ex}
\end{figure}

It is critical yet challenging to show detailed hand appearance with consistency in action-specific motion in hand-centric video generation: the hands are typically relatively small in the pixel-space, the hands are articulated, the motion is intricate, and the background is often cluttered.
We hence propose a new Hand Refinement Loss (HRL) function $\mathcal{L}_{HR}$ that improves the subtle but critical details, emphasizing the hands (see Fig. \ref{fig:handpose}). Our approach is more lightweight (applied as a loss function) than previous hand generation work which require a separate model or heavy post processing~\cite{narasimhaswamy2024handiffuser, lu2024handrefiner, yang2024annotated, fu2025handrawerleveragingspatialinformation} \TODO{May need to add these methods as baselines}. %

Let us define the representation of the hand.  In a given frame $l$, let $P_{l} \in \mathds{R}^{j \times 2} \;(j \in \{0, \dots J\})$ represent the hand structure, containing the 2D coordinates of the $J$ joints. 
In this work, we assume there are at most two hands in one frame, aligned with popular hand-centric activity datasets (e.g., EpicKitchens and Ego4D~\cite{damen2022rescaling, grauman2022ego4d}). We follow a commonly used hand pose skeleton topology~\cite{lugaresi2019mediapipe} which defines 21 joints to represent structure of one hand. Therefore, $J$ is set to 42 in all our experiments.

We compute the Hand Refinement Loss between the generated video $V^{\text{gen}} \in \mathds{R}^{L\,\boxtimes\,3}$ from stage 2 and the input training video $V^{\text{train}} \in \mathds{R}^{L\,\boxtimes\,3}$, both not including the input context image.  
To produce $P^{\text{gen}}_{l}$ and $P^{\text{train}}_{l}$, we first apply the SOTA hand pose detector $\Upsilon$ on both videos~\cite{lugaresi2019mediapipe}. 
$\Upsilon$ is aware of the temporal information in the video and configured to produce joint coordinates for left and right hands. 
We normalize the joints detected by $\Upsilon$ in terms of frame size to $[0, 1]$. If some joints in the skeleton topology are not detected by $\Upsilon$ (i.e., the number of detected joints is less than $J$), we set the coordinates of the missing joints in either $P^{\cdot}_{l}$ as 0. 
Those missing joints hence do not contribute to model optimization for this frame, but the same computational effiency is maintained.

The point of adding the hand refinement loss is to encourage the generation of realistic, articulated hand pose for the specified activity.  We hence freeze the hand detector $\Upsilon$ and optimize the stage 2 noise predictor to encourage better hand rendering, as follows.
Given the hand pose sequences in the training and generated videos, we compute the Mean Square Error (MSE) between them as the loss value:
\begin{align}
    \mathcal{L}_{HR} = 
    \frac{1}{L} \sum_{l=1}^{L} \frac{1}{J} 
    \left\| P_{l}^{\text{gen}} - P_{l}^{\text{train}} \right\|^2_{F}\quad.
    \label{eq:hand_loss}
\end{align}
Note that $\mathcal{L}_{HR}$ starts to contribute densely at the latter iterations of the training when the rough hand structure in the generated video can be detected by $\Upsilon$. 
Clearly, computing $\mathcal{L}_{HR}$ requires the decoded video $V^{\text{gen}}$ available; we compute this video at each training iteration directly from the denoised latent video $z'_0$ (see \S\ref{sec:fullloss}, Eq. (\ref{eq:zprimenaut})).

\begin{figure*}
    \centering
    \includegraphics[width=1\linewidth]{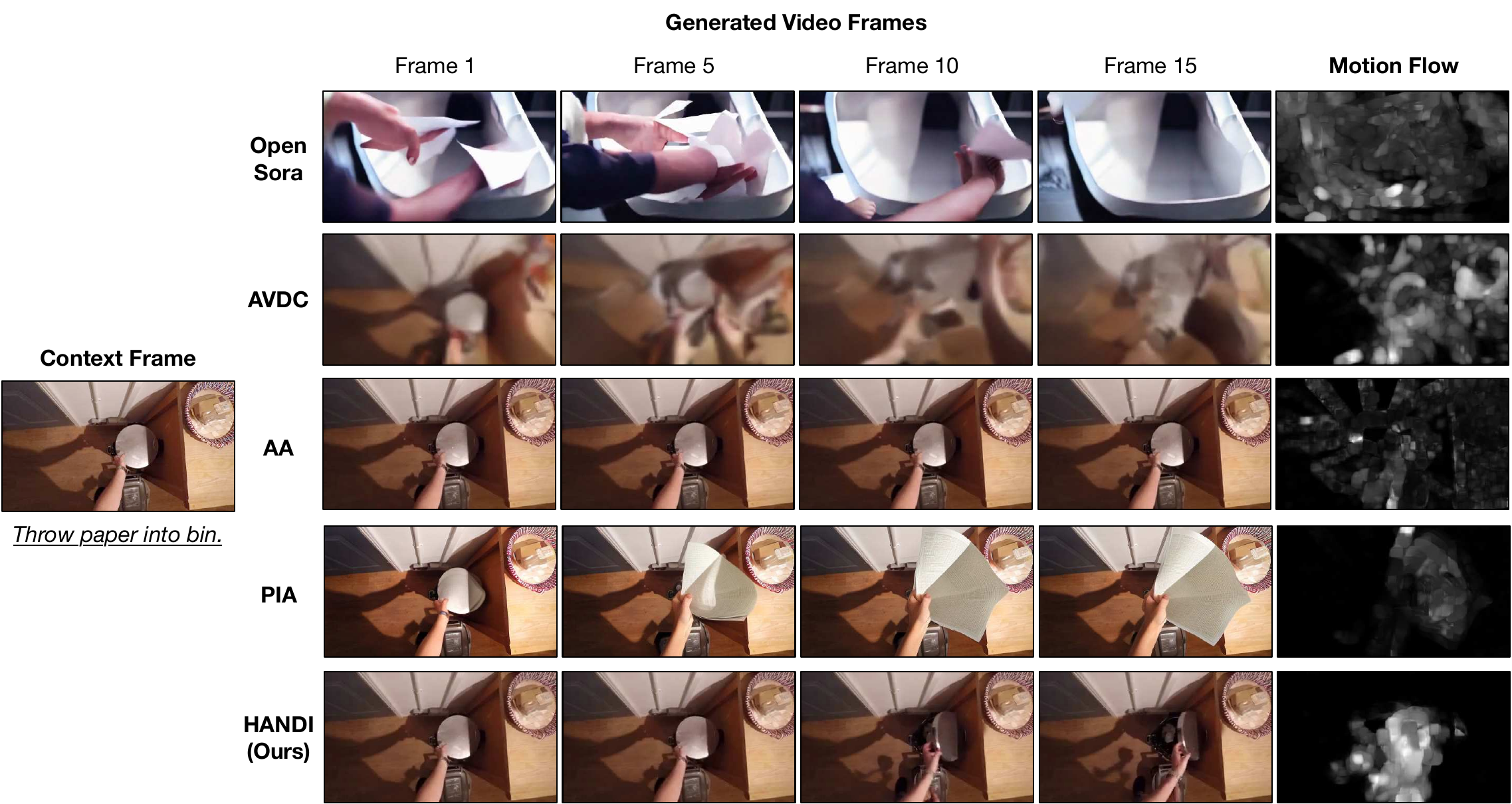}
    \vspace{-4ex}
    \caption{Qualitative results comparing with baselines. Our method generates videos that contain subtle hand motion that corresponds to the action description. This is because our method focuses on the effective motion region instead of cluttered background and our method is optimized with a tailored Hand Refinement Loss. The motion flow of the generated video provides more straightforward visualization and yield the same conclusion. Please see more video results in supplementary materials. \TODO{(Supp) a row of zoomed-in results} \TODO{(Supp and this figure) generated masks.} \TODO{(Supp) update CogVideoX qualitative results with more iterations of fine-tuning.}}
    \label{fig:mainresults}
    \vspace{-3ex}
\end{figure*}

\subsection{Latent- and Pixel-Space Loss Portfolio}
\label{sec:fullloss}

With our innovations in automatically generating the MA mask and in the hand refinement, we require loss computation both in the latent and the pixel space at each stage.  

\noindent\textbf{Latent Space}\quad We define our the noise prediction loss $\mathcal{L}_{noise}$ as follows, reflecting the conditions from input video, text, and MA mask for HCVG:
\begin{equation}
    \small
    \mathbf{\mathcal{L}_{noise}} = \mathbb{E}_{(V, E_{text}, M_{\{prior, gen\}}), \epsilon \sim \mathcal{N}(0, 1), t} \left[ \left\| \epsilon - \epsilon_{\theta}(z_t, t) \right\|^2_2 \right]
\end{equation}
where $V$ is the training video (including the context image~$I$), $E_{text}$ is the feature representation of action prompt, $M_{\{prior, gen\}}$ is the prior or generated MA mask for stage 1 or 2 respectively,  $\epsilon$ is the sampled Gaussian noise and $\epsilon_{\theta}$ is the noise predicted by the noise predictor $D_{\{1, 2\}}$ conditioned on the noise-added latent video $z_t$ with the randomly sampled noise step $t$. 

\noindent\textbf{Pixel Space}\quad For the pixel-level loss $\mathcal{L}_{mIoU}$ and $\mathcal{L}_{HR}$, we first decode the denoised latent video $z'_0$ to the pixel space and then compute the loss between the generated video and the target video. 
We obtain the clean latent video $z'_{0}$ by the following equation:
\begin{align}
    z'_0 = \frac{z_t - \sqrt{1 - \prod_{i=1}^t (1 - \beta_i)} \, \epsilon}{\sqrt{\prod_{i=1}^t (1 - \beta_i)} \,}.
    \label{eq:zprimenaut}
\end{align}
Then, we decode the actual video $V^{\text{gen}}$ from $z'_0$, using the same process we described above during inference.

\noindent\textbf{Stage 1 Loss}\quad We apply a mIoU loss $\mathcal{L}_{mIoU}$ in pixel space in addition to $\mathcal{L}_{noise}$ to have the final loss:
\begin{equation}
    \mathcal{L}_{stage1} = \mathcal{L}_{noise} + \alpha \mathcal{L}_{mIoU}
\end{equation}
where $\alpha$ is a hyperparameter tuned empirically. The Mean Intersection over Union (mIoU) loss is defined as
\begin{equation}
\footnotesize
\mathcal{L}_{\text{mIoU}} = 1 - \frac{1}{L} \sum_{l=1}^{L} \frac{\sum_{i,j} \left( M_{l}^{\text{gen}} \cdot M_{l}^{\text{train}} \right)}{\sum_{i,j} \left( M_{l}^{\text{gen}} + M_{l}^{\text{train}} - M_{l}^{\text{gen}} \cdot M_{l}^{\text{train}} \right)},
\end{equation}
where \( M_{l}^{\text{gen}} \) and \( M_{l}^{\text{train}} \) are the generated and training binary masks for frame \( l \), summed over spatial dimensions \( i \) and \( j \). \( L \) denotes the number frames in the output video.

\noindent\textbf{Stage 2 Loss}\quad We apply the Hand Refinement Loss $\mathcal{L}_{HR}$ in addition to $\mathcal{L}_{noise}$ to have the final loss:
\begin{equation}
    \mathcal{L}_{stage2} = \mathcal{L}_{noise} + \eta \mathcal{L}_{HR}\quad ,
\end{equation}
where the coefficient $\eta$ is a hyperparameter.

\section{Experiments}
\label{sec:exp}

\noindent\textbf{Datasets}\quad 
We evaluate on two raw hand-action videos corpora: Epic-Kitchens-100 (EK)~\cite{damen2022rescaling} and Ego4D~\cite{grauman2022ego4d}, using LEGO’s quality-checked splits~\cite{lai2023lego}.  
EK has 60,429/8,857 train/test samples (2.4s avg.), and Ego4D has 85,521/9,931 (1.07s avg.).  
These datasets reflect real-world challenges: clutter, lighting variation, and small hand regions—spanning 97/1772 verbs, 300/4336 nouns, and 45/74 kitchen or location settings in EK/Ego4D.

\begin{table*}[th!]
    \centering
    \footnotesize
    
    \begin{tabular}{
    ccccccccccc} 
        \toprule
        \multirow{2}{*}{\textbf{Benchmark}} & 
        \multirow{2}{*}{\textbf{Method}} & 
        \multirow{2}{*}{\textbf{HS-Err.} $\downarrow$} & 
        \multicolumn{2}{c}{\textbf{VisualSim.-Frame}} & 
        \multicolumn{2}{c}{\textbf{VisualSim.-Video}} & 
        \multicolumn{1}{c}{\textbf{Consistency}} & 
        \multicolumn{2}{c}{\textbf{SemanticSim.}} 
        \\
        \cmidrule(lr){4-5} \cmidrule(lr){6-7} 
        \cmidrule(lr){9-10}
        & 
        & 
        & \textbf{FID $\downarrow$} & \textbf{$\text{CLIP}_{GT}$ $\uparrow$} & \textbf{FVD $\downarrow$} & \textbf{EgoVLP $\uparrow$} & \textbf{$\text{CLIP}_{Cs.}$ $\uparrow$} & \textbf{$\text{CLIP}_{Tx.}$ $\uparrow$} & \textbf{BLIP $\uparrow$} 
        \\
        
        \midrule
        \multirow{8}{*}{\textbf{EpicKitchens}} 
        & LFDM~\cite{ni2023conditional} & 0.01987 & 39.37 & 0.9241 & 129.80 & 0.354 & 0.9826 & 28.37 & 0.235\\
        & AA~\cite{dai2023fine} & 0.01908 & \underline{5.49} & \underline{0.9588} & 171.29 & 0.338 & 0.9843 & 29.97 & \underline{0.295}\\
        & AVDC~\cite{du2023learning} & 0.01969 & 140.34 & 0.8918 & \textbf{81.39} & 0.197 & 0.9582 & 24.66 & 0.116 \\
        & PIA~\cite{zhang2024pia} & 0.01826 & 24.70 & 0.9446 & 212.88 & \underline{0.361} & \underline{0.9849} & \underline{30.06} & 0.294\\
        & Open Sora~\cite{opensora} & 0.01968 & 135.34 & 93.823 & 124.52 & 0.187 & 0.9573 & 24.46 & 0.186\\
        & DynamiCrafter~\cite{xing2023dynamicrafter} & \underline{0.01716} & 43.56 & 0.9306 & 175.79 & 0.348 & 0.9131 & 28.22 & 0.288\\
        & CogVideoX~\cite{yang2024cogvideox} & 0.01981 & 127.51 & 0.9362 & 121.06 & 0.214 & 0.9677 & 25.13 & 0.176\\
        & \cellcolor{gray!20} HANDI & \cellcolor{gray!20} \textbf{0.01512} & \cellcolor{gray!20} \textbf{5.27} & \cellcolor{gray!20} \textbf{0.9590} & \cellcolor{gray!20} \underline{101.89} & \cellcolor{gray!20} \textbf{0.377} & \cellcolor{gray!20} \textbf{0.9896} & \cellcolor{gray!20} \textbf{31.14} & \cellcolor{gray!20} \textbf{0.298}\\
        \midrule
        \multirow{6}{*}{\textbf{Ego4D}} 
        & LFDM~\cite{ni2023conditional} & 0.02127 & 50.67 & 0.9204 & 126.71 & 0.535 & 0.9821 & 26.93 & 0.221 \\
        & AA~\cite{dai2023fine} & 0.02393 & \underline{21.83} & \underline{0.9647} & 129.60 & \underline{0.642} & \underline{0.9894} & 28.56 & \underline{0.260}\\
        & AVDC~\cite{du2023learning} & \underline{0.02117} & 144.91 & 0.8816 & 107.82 & 0.261 & 0.9722 & 24.17 & 0.155\\
        & PIA~\cite{zhang2024pia} & 0.02393 & 34.62 & 0.9457 & \underline{104.38} & 0.603 & 0.9746 & \textbf{29.15} & 0.219\\
        & Open Sora~\cite{opensora} & 0.02142 & 141.90 & 0.8716 & 117.87 & 0.252 & 0.9753 & 24.12 & 0.172\\
        & DynamiCrafter~\cite{xing2023dynamicrafter} & 0.02203 & 57.21 & 0.9386 & 181.24 & 0.336 & 0.9489 & 26.67 & 0.231\\
        & CogVideoX~\cite{yang2024cogvideox} & 0.03079 & 187.20 & 0.8962 & 165.18 & 0.201 & \textbf{0.9900} & 28.22 & 0.147\\
        & \cellcolor{gray!20} HANDI & \cellcolor{gray!20} \textbf{0.01939} & \cellcolor{gray!20} \textbf{21.51} & \cellcolor{gray!20} \textbf{0.9651} & \cellcolor{gray!20} \textbf{103.15} & \cellcolor{gray!20} \textbf{0.664} & \cellcolor{gray!20} 0.9873 & \cellcolor{gray!20} \underline{28.63} & \cellcolor{gray!20} \textbf{0.263}\\
        \midrule
        \multirow{5}{*}{\textbf{Motion Intensive}} 
        & LFDM~\cite{ni2023conditional} & 0.02053 & 56.95 & 0.9254 & 137.44 & 0.303 & 0.9825 & 28.39 & 0.210\\
        & AA~\cite{dai2023fine} & \underline{0.01764} & \underline{23.93} & \textbf{0.9591} & 115.14 & \underline{0.368} & \underline{0.9845} & 30.07 & 0.276\\
        & AVDC~\cite{du2023learning} & 0.02143 & 148.11 & 0.8933 & \textbf{85.97} & 0.204 & 0.9579 & 24.60 & 0.102\\
        & PIA~\cite{zhang2024pia} & 0.01940 & 40.97 & 0.9448 & 217.59 & 0.330 & 0.9719 & \underline{30.09} & \underline{0.280}\\
        & \cellcolor{gray!20} HANDI & \cellcolor{gray!20} \textbf{0.01663} & \cellcolor{gray!20} \textbf{23.79} & \cellcolor{gray!20} \underline{0.9589} & \cellcolor{gray!20} \underline{114.52} & \cellcolor{gray!20} \textbf{0.371} & \cellcolor{gray!20} \textbf{0.9849} & \cellcolor{gray!20} \textbf{31.12} & \cellcolor{gray!20} \textbf{0.327}\\
        \bottomrule
    \end{tabular}
    \vspace{-2ex}
    \caption{Quantitative results on EpicKitchens~\cite{damen2022rescaling}, Ego4D~\cite{grauman2022ego4d} and a Motion Intensive subset of EpicKitchens. \colorbox{gray!20}{HANDI (ours)} outperforms all baselines across all metrics on at least one benchmark. VisualSim.-Frame represents the aspect of visual similarity at frame level; VisualSim.-Video represents visual similarity at video level; SemanticSim. stands for Semantic Similarity. For each column in each benchmark section, \textbf{bold} represents the best performance and \underline{underline} stands for the second best one. 
    \TODO{Update CogVideoX results with more iterations of fine-tuning.}}
    \label{tab:main}
    \vspace{-2ex}
\end{table*}

\noindent\textbf{Metrics}\quad
We conduct a comprehensive quantitative evaluation.
For \emph{visual similarity} between generated and ground truth videos, we 
use the  image-to-image Fréchet Inception Distance (FID)~\cite{heusel2017gans} and $\text{CLIP}_{GT}$ score%
~\cite{radford2021learning} as our averaged frame-level similarity, and the Fréchet Video Distance (FVD)~\cite{unterthiner2019fvd} and EgoVLP score~\cite{lai2023lego, lin2022egocentric} at the video level. 
For \emph{semantic similarity} between the prompt and the generated video, we adopt the frame-level $\text{CLIP}_{Tx.}$ score~\cite{wu2024towards} and the video level BLIP score~\cite{lai2023lego, li2022blip}. 
We measure \emph{motion coherency} of the generated video. We apply the Consistency CLIP score $\text{CLIP}_{Cs.}$, an average CLIP similarity between consecutive frames~\cite{radford2021learning}. FVD also reflects temporal coherency. 
In the end, to evaluate the \emph{generated hand quality}, we report Hand Structure Error (HS-Err) which is defined as in Eq. (\ref{eq:hand_loss}).
We report all metrics on $L$ frames in the videos not including the context image $I$.

\noindent\textbf{Implementation Details}\quad For each video, we evenly sample 16 frames where the first frame is the context image $I$ and the remaining 15 frames (i.e., $L=15$) comprise the video to be denoised during training. The VAE~\cite{rombach2022high}, instruction tuning module~\cite{rombach2022high}, text encoder~\cite{rombach2022high} and hand joints detection model~\cite{lugaresi2019mediapipe} are previous work and frozen. The noise predictors are initialized with pretrained weights~\cite{dai2023fine} and optimized during training. The coefficients in the loss functions are $\alpha=0.1$ and $\eta=0.1$. We train our model for 50 epochs with batch size 48 and learning rate $5 \times 10^{-5}$. These hyperparameters are chosen empirically on small subsets of the training data. We use mixed precision with bf16 on 6 $\times$ H100 GPUs.

\begin{table*}[h!]
    \centering
    \footnotesize
    \begin{tabular}{c
    ccccccccccccc
    }
        \toprule
        \multirow{2}{*}{\textbf{Scope}} & \multirow{2}{*}{\textbf{No.}} & \multirow{2}{*}{\textbf{Mask}} & \multirow{2}{*}{\textbf{HRL}} &         
        \multirow{2}{*}{\textbf{HS-Err.} $\downarrow$} & 
        \multicolumn{2}{c}{\textbf{VisualSim.-Frame}} & 
        \multicolumn{2}{c}{\textbf{VisualSim.-Video}} & 
        \multicolumn{1}{c}{\textbf{Consistency}} & 
        \multicolumn{2}{c}{\textbf{SemanticSim.}} 
        \\
        \cmidrule(lr){6-7} \cmidrule(lr){8-9} \cmidrule(lr){11-12} 
        & & & 
        & & \textbf{FID $\downarrow$} & \textbf{$\text{CLIP}_{GT}$ $\uparrow$} & \textbf{FVD $\downarrow$} & \textbf{EgoVLP $\uparrow$} & \textbf{$\text{CLIP}_{Cs.}$ $\uparrow$} & \textbf{$\text{CLIP}_{Tx.}$ $\uparrow$} & \textbf{BLIP $\uparrow$} 
        \\
        \midrule
        \multirow{7}{*}{\textbf{MA}} 
        & 1 & $\times$ & $\times$ & 0.02067 & 13.516 & 0.9551 & 172.56 & 0.211 & 0.9677 & 27.862 & 0.261\\
        & 2 & Stage 1 & $\times$ & 0.01949 & 13.023 & 0.9549 & 122.31 & 0.214 & 0.9650 & 26.840 & 0.271\\
        & 3 & $\times$ & \checkmark & 0.01771 & 13.440 & 0.9549 & 169.99 & 0.339 & \underline{0.9780} & 27.863 & 0.269\\
        & \cellcolor{gray!20}4 & \cellcolor{gray!20} Stage 1 & \cellcolor{gray!20} \checkmark & \cellcolor{gray!20} \textbf{0.01504} & \cellcolor{gray!20} \underline{10.692} & \cellcolor{gray!20} \textbf{0.9563} & \cellcolor{gray!20} \textbf{100.52} & \cellcolor{gray!20}\textbf{0.390} & \cellcolor{gray!20} 0.9753 & \cellcolor{gray!20} \underline{28.041} & \cellcolor{gray!20} \textbf{0.283}\\
        & 5 & Prior & \checkmark & \underline{0.01701} & 12.530 & \underline{0.9551} & 132.19 & 0.355 & \textbf{0.9848} & 27.909 & 0.275\\
        & 6 & Prior & $\times$ & 0.02149 & 11.335 & 0.9539 & 116.09 & 0.352 & 0.9714 & 28.011 & 0.262\\
        \cmidrule(lr){2-12} 
        & 7 & GT & $\times$ & 0.01804 & 10.847 & 0.9547 & \underline{101.53} & 0.367 & 0.9755 & 28.023 & 0.272 \\
        & 8 & GT & $\checkmark$ & 0.01762 & \textbf{9.816} & 0.9536 & 114.44 & \underline{0.372} & 0.9631 & \textbf{28.102} & \underline{0.281} \\
        \midrule
        \midrule
        \multirow{7}{*}{\textbf{Full}} 
        & 1 & $\times$ & $\times$ & 0.01908 & 5.497 & 0.9588 & 171.29 & 0.338 & 0.9843 & 29.970 & 0.295\\
        & 2 & Stage 1 & $\times$ & 0.01940 & 5.570 & \underline{0.9589} & 124.42 & 0.197 & 0.9881 & 29.949 & 0.296\\
        & 3 & $\times$ & \checkmark & 0.01760 & 5.384 & \underline{0.9589} & 168.85 & 0.338 & 0.9889 & 29.929 & 0.294\\
        & \cellcolor{gray!20}4 & \cellcolor{gray!20} Stage 1 & \cellcolor{gray!20} \checkmark & \cellcolor{gray!20} \textbf{0.01512} & \cellcolor{gray!20} \underline{5.272} & \cellcolor{gray!20} \textbf{0.9590} & \cellcolor{gray!20} \textbf{101.89} & \cellcolor{gray!20} 0.377 & \cellcolor{gray!20} \textbf{0.9896} & \cellcolor{gray!20} \underline{30.055} & \cellcolor{gray!20}\textbf{0.298}\\
        & 5 & Prior & \checkmark & 0.01670 & 5.285 & 0.9588 & 137.01 & 0.354 & \underline{0.9869} & 30.045 & \underline{0.297}\\
        & 6 & Prior & $\times$ & 0.02146 & 5.613 & 0.9589 & 161.48 & 0.341 & 0.9748 & 30.025 & 0.296\\
        \cmidrule(lr){2-12} 
        & 7 & GT & $\times$ & 0.01935 & 5.287 & \underline{0.9589} & \underline{102.43} & \underline{0.380} & 0.9842 & 30.050 & \underline{0.297} \\
        & 8 & GT & $\checkmark$ & 0.01738 & \textbf{5.194} & 0.9587 & 113.57 & \textbf{0.382} & 0.9748 & \textbf{30.096} & 0.295 \\
        \bottomrule
    \end{tabular}
    \vspace{-2ex}
    \caption{Ablation study on the EpicKitchens benchmark~\cite{damen2022rescaling}, examining the impact of Motion Area (MA) mask generation (``Mask") and Hand Refinement Loss (``HRL") on model performance. 
   \colorbox{gray!20}{The highlighted row} is our full HANDI method. When Mask is ``GT'', we use the ground truth mask which is supposed to be the upper bound of the performance but we still outperform it in more than half of the metrics.}
    \label{tab:ablation}
    \vspace{-2ex}
\end{table*}

\subsection{Comparisons with the State-of-the-Art}
\label{sec:sotacomparison}
\label{sec:sota}

\noindent\textbf{Baselines}\quad
\TODO{Fight for the credits on fine-tuning all the baselines}
Our baseline methods include TI2V models AnimateAnything~\cite{dai2023fine}, PIA~\cite{zhang2024pia}, DynamiCrafter~\cite{xing2023dynamicrafter}, AVDC~\cite{du2023learning} and LFDM~\cite{ni2023conditional}, as well as T2V models CogVideoX (5B)~\cite{yang2024cogvideox}, and Open Sora~\cite{opensora}. For space, we describe only some of these here. 
PIA integrates temporal alignment and conditioning layers but does not explicitly target hand precision in cluttered environments. 
DynamiCrafter adapts a video diffusion prior for animating still images by injecting text-aligned context representations. However, its reliance on implicit motion priors makes it susceptible to generating unnatural and extended dynamics when fine-grained hand-object interactions are required. %
For the T2V models, CogVideoX and Open Sora focus on generating long-form, high-motion videos but lack direct image conditioning, making them ineffective for HCVG as our results show. CogVideoX employs a 3D Variational Autoencoder and an expert transformer to improve temporal consistency but suffers from flickering artifacts across frames. 
Since none of the baselines are originally tailored for HCVG, we fine-tune all models on the Epic-Kitchens and Ego4D benchmarks for 50 epochs (200,000 iterations, batch size 16) for fair evaluation. %

\noindent\textbf{Quantitative and Qualitative Results.}
Table~\ref{tab:main} details our compelling quantitative comparisons. 
HANDI performs best across nearly \emph{all} metrics and SOTA baselines,
including the motion intensive samples---the subset of EK that have top 10\% largest $M_{video}$ (\S\ref{sec:rom}). 
In a few cases (4 values among 24), strong baselines (AVDC and AA) perform best; yet, neither shows consistent advantage across \emph{any} metrics on all benchmarks. For example, although AVDC achieves the lowest FVD in EpicKitchens (the full set and motion-intensive subsets), it ranks the third for FVD on Ego4D, which is a more challenging benchmark with more diverse actions, objects, and environments (i.e., outside of the kitchen~\cite{grauman2022ego4d}). 
Notably, HANDI also produces the most semantically relevant videos, reflected in the highest $\text{CLIP}_{Tx.}$ and BLIP scores, which measure alignment with the text prompt. 
We discuss the (strong performance in the) hand structure error below in \S\ref{sec:ablation}.

We show qualitative samples in Fig.~\ref{fig:mainresults} (and videos in supplementary materials). The key limitations of prior TI2V methods are evident, including those performing well quantitatively (PIA, AVDC, AA). Despite their image-conditioning capabilities, these baselines suffer from unwanted background artifacts or blurry/rippled scene effects due to the lack of focus on the Motion Area. 
For example, 
AVDC generates blurry videos because it is designed for robot policy learning with low-resolution video. 
On the other hand, the hand quality generated baselines are rough or the motion is unnatural even if not affected by the scene effects (e.g, PIA and AA). These shortcomings stem from their lack of explicit hand-motion constraints, leading to minor flickering or unnatural hand movements. 
Likewise, the T2V baselines (Open Sora and CogVideoX) struggle in the HCVG setting, as expected with no image context. 

Fig.~\ref{fig:mainresults} also shows how HANDI successfully localizes motion to the hand interaction region, avoiding unnecessary background changes and global effects. Motion flow analysis further corroborates this observation, showing that HANDI concentrates motion within the targeted region while competing methods exhibit undesired movement in surrounding areas.

\noindent\textbf{Computational Efficiency Analysis.} Although \textsc{HANDI} introduces a separate stage for motion–area prediction, its overall run-time is on par with (or better than) comparable single-stage systems.  
On EpicKitchens, synthesising a \(256\times256\) clip on one NVIDIA H100 takes \textbf{8.6\,s}.  
For reference, 
the light but less expressive 
AnimateAnything completes in 5.7\,s\;\cite{dai2023fine}, DynamiCrafter in 9.3\,s\;\cite{xing2023dynamicrafter}, PIA in 13.8\,s\;\cite{zhang2024pia}, OpenSora in 44\,s\;\cite{opensora}, AVDC in 66\,s\;\cite{du2023learning}, and CogVideoX in 108\,s\;\cite{yang2024cogvideox}.  
\textsc{HANDI} therefore delivers its hand-fidelity gains without imposing a meaningful latency penalty.

\subsection{Ablation Study}
\label{sec:ablation}
\begin{figure}
    \centering
    \includegraphics[width=1\linewidth]{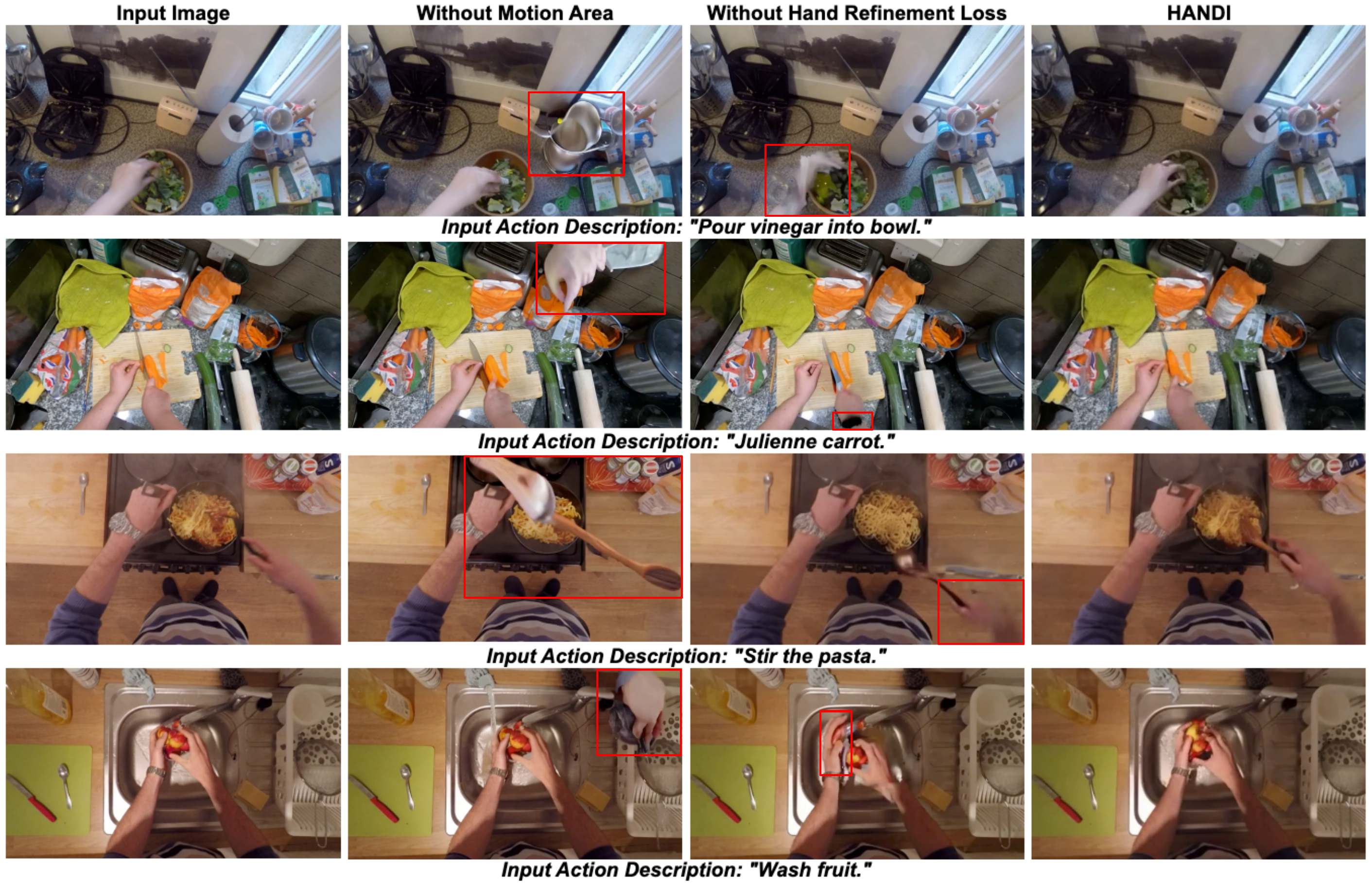}
    \vspace{-2ex}
    \caption{The visualization for ablation study showing the independent effectiveness of Motion Area generation and Hand Refinement Loss. The \fcolorbox{red}{white}{red boxes} are not in the generated videos. They are drawn for better illustration, showing the area that reflects the effectiveness of the ablated components. \TODO{a row of zoomed-in results}}
    \label{fig:ablation}
    \vspace{-2ex}
\end{figure}

We conduct an ablation study to investigate the effect of our two new model elements---generated Motion Area (MA) and Hand Refinement Loss (HRL); shown in~\cref{tab:ablation} and~\cref{fig:ablation}. 
In the ablation, we not only assess performance on the full frames (Scope is ``Full'') but also assess it on the ground truth motion area $M_{video}$ (\S\ref{sec:rom}) since this is the detailed hand-centric part of the video we focus on in HCGV (Scope is ``MA'').  
Note that no mask (Mask is $\times$) means the motion area is set to the full frame for stage 2 generation.

Comparing rows 1 and 3 in both scopes assesses the impact of the model generated motion area.  Clearly, allowing the system to generate a motion area in which to focus the stage 2 diffusion leads to higher quality video while preventing the generation of distractions and hallucinations in the background (as shown in~\cref{fig:ablation}).
Rows 2 and 3 compare the impact of the HRL.  
As shown in~\cref{fig:ablation} (rightmost two columns), without HRL, the generated hand could be blurry (fig. row 1 and 3) or cut off (fig. row 2) or distorted (fig. row 4). This demonstrates the capability of the HRL for detailed hand appearance improvement.   
Both sets of quantitative comparisons show significant improvements
when the scope is set to the motion area, indicating better quality of the generated hand details, especially for FID and FVD, which capture the perceptual quality of the generated video.

\begin{TODOB}
\textbf{Is only using Stage 1 helpful?} \{MA, Full\}-1 vs. 2. On MA, we loose on $\text{CLIP}_{GT}$, $\text{CLIP}_{Cs.}$, $\text{CLIP}_{Tx.}$. This reflects the limitation of using CLIP to measure video generation performance. On Full, we lose HS, FID, CLIP Tx,. 

\textbf{Comparing with Prior mask.} With HS (4 vs. 5), on MA, we lose on CLIP Cs. On Full, we win on all. Without HS (2 vs. 6), on MA, we lose FID, EgoVLP, CLIP cs, CLIP Tx. On Full, we lose FID, CLIP GT (equ), EgoVLP, CLIP Tx, BLIP (eq). With HS (3 vs. 4), on MA, we lose CLIP Cs. On Full, we win on all. 

\textbf{Comparing with GT.} With HS (4 vs. 8), on MA, we win HS, CLIP GT, FVD, EgoVLP. On Full, we win HS, CLIP GT, FVD, CLIP Cs., BLIP. 

\textbf{Conclusion.} HRL not only improves performance independently but also is a necessary component to unlock the effectiveness of Stage 1. 
\end{TODOB}

\section{Conclusion}
We tackle the challenges in generating videos of hands performing goal-oriented actions, an understudied but important aspect of video generation for applications like human and robot learning. These scenarios demand high-fidelity hand motion, conditioned on a single image and text prompt. Our method, HANDI, builds on diffusion-based video generation with two innovations: automatic localisation of the motion area, avoiding background hallucination, and a hand refinement loss that encourages anatomically realistic and temporal consistent hand motion. Our comparisons on real-world datasets demonstrate consistent gains over prior methods while maintaining efficiency.

{
    \small
    \bibliographystyle{ieeenat_fullname}
    \bibliography{main}

\begin{thebibliography}{82}
\providecommand{\natexlab}[1]{#1}
\providecommand{\url}[1]{\texttt{#1}}
\expandafter\ifx\csname urlstyle\endcsname\relax
  \providecommand{\doi}[1]{doi: #1}\else
  \providecommand{\doi}{doi: \begingroup \urlstyle{rm}\Url}\fi

\bibitem[Aifanti et~al.(2010)Aifanti, Papachristou, and Delopoulos]{aifanti2010mug}
Niki Aifanti, Christos Papachristou, and Anastasios Delopoulos.
\newblock The mug facial expression database.
\newblock In \emph{11th International Workshop on Image Analysis for Multimedia Interactive Services WIAMIS 10}, pages 1--4. IEEE, 2010.

\bibitem[Ashutosh et~al.(2024)Ashutosh, Xue, Nagarajan, and Grauman]{ashutosh2024detours}
Kumar Ashutosh, Zihui Xue, Tushar Nagarajan, and Kristen Grauman.
\newblock Detours for navigating instructional videos.
\newblock In \emph{Proceedings of the IEEE/CVF Conference on Computer Vision and Pattern Recognition}, pages 18804--18815, 2024.

\bibitem[Bain et~al.(2021)Bain, Nagrani, Varol, and Zisserman]{bain2021frozen}
Max Bain, Arsha Nagrani, G{\"u}l Varol, and Andrew Zisserman.
\newblock Frozen in time: A joint video and image encoder for end-to-end retrieval.
\newblock In \emph{Proceedings of the IEEE/CVF international conference on computer vision}, pages 1728--1738, 2021.

\bibitem[Bakr et~al.(2023)Bakr, Sun, Shen, Khan, Li, and Elhoseiny]{bakr2023hrs}
Eslam~Mohamed Bakr, Pengzhan Sun, Xiaoqian Shen, Faizan~Farooq Khan, Li~Erran Li, and Mohamed Elhoseiny.
\newblock Hrs-bench: Holistic, reliable and scalable benchmark for text-to-image models.
\newblock In \emph{Proceedings of the IEEE/CVF International Conference on Computer Vision}, pages 20041--20053, 2023.

\bibitem[Basu et~al.(2023)Basu, Saberi, Bhardwaj, Chegini, Massiceti, Sanjabi, Hu, and Feizi]{basu2023editval}
Samyadeep Basu, Mehrdad Saberi, Shweta Bhardwaj, Atoosa~Malemir Chegini, Daniela Massiceti, Maziar Sanjabi, Shell~Xu Hu, and Soheil Feizi.
\newblock Editval: Benchmarking diffusion based text-guided image editing methods.
\newblock \emph{arXiv preprint arXiv:2310.02426}, 2023.

\bibitem[Brooks et~al.(2023)Brooks, Holynski, and Efros]{brooks2023pix2pix}
T. Brooks, A. Holynski, and A.~A. Efros.
\newblock Instructpix2pix: Learning to follow image editing instructions.
\newblock In \emph{Proceedings of IEEE/CVF Conference on Computer Vision and Pattern Recognition}, 2023.

\bibitem[Cen et~al.(2024)Cen, Pi, Peng, Shen, Yang, Shuai, Bao, and Zhou]{cen2024text_scene_motion}
Zhi Cen, Huaijin Pi, Sida Peng, Zehong Shen, Minghui Yang, Zhu Shuai, Hujun Bao, and Xiaowei Zhou.
\newblock Generating human motion in 3d scenes from text descriptions.
\newblock In \emph{CVPR}, 2024.

\bibitem[Ceylan et~al.(2023)Ceylan, Huang, and Mitra]{ceylan2023pix2video}
D. Ceylan, C.-H.~P. Huang, and N.~J. Mitra.
\newblock Pix2video: Video editing using image diffusion.
\newblock In \emph{Proceedings of the IEEE/CVF International Conference on Computer Vision (ICCV)}, 2023.

\bibitem[Chandler and Sweller(1991)]{chandler1991cognitive}
Paul Chandler and John Sweller.
\newblock Cognitive load theory and the format of instruction.
\newblock \emph{Cognition and instruction}, 8\penalty0 (4):\penalty0 293--332, 1991.

\bibitem[Chane-Sane et~al.(2023)Chane-Sane, Schmid, and Laptev]{chanesane2023vip}
Elliot Chane-Sane, Cordelia Schmid, and Ivan Laptev.
\newblock Learning video-conditioned policies for unseen manipulation tasks.
\newblock In \emph{ICRA}, 2023.

\bibitem[{Cheikh Youssef} et~al.(2022){Cheikh Youssef}, Aydin, Canning, Khan, Ahmen, and Dasgupta]{YoAyCaSI2022}
S. {Cheikh Youssef}, A. Aydin, A. Canning, N. Khan, K. Ahmen, and P. Dasgupta.
\newblock Learning surgical skills through video-based education: A systematic review.
\newblock \emph{Surgical Innovation}, 30\penalty0 (2):\penalty0 220--238, 2022.

\bibitem[Chen et~al.(2024{\natexlab{a}})Chen, Zhang, Cun, Xia, Wang, Weng, and Shan]{chen2024videocrafter2}
Haoxin Chen, Yong Zhang, Xiaodong Cun, Menghan Xia, Xintao Wang, Chao Weng, and Ying Shan.
\newblock Videocrafter2: Overcoming data limitations for high-quality video diffusion models, 2024{\natexlab{a}}.

\bibitem[Chen et~al.(2024{\natexlab{b}})Chen, Siarohin, Menapace, Deyneka, Chao, Jeon, Fang, Lee, Ren, Yang, et~al.]{chen2024panda}
Tsai-Shien Chen, Aliaksandr Siarohin, Willi Menapace, Ekaterina Deyneka, Hsiang-wei Chao, Byung~Eun Jeon, Yuwei Fang, Hsin-Ying Lee, Jian Ren, Ming-Hsuan Yang, et~al.
\newblock Panda-70m: Captioning 70m videos with multiple cross-modality teachers.
\newblock In \emph{Proceedings of the IEEE/CVF Conference on Computer Vision and Pattern Recognition}, pages 13320--13331, 2024{\natexlab{b}}.

\bibitem[Dai et~al.(2023)Dai, Zhang, Yao, Qiu, Zhu, Qin, and Wang]{dai2023fine}
Zuozhuo Dai, Zhenghao Zhang, Yao Yao, Bingxue Qiu, Siyu Zhu, Long Qin, and Weizhi Wang.
\newblock Fine-grained open domain image animation with motion guidance.
\newblock \emph{arXiv preprint arXiv:2311.12886}, 2023.

\bibitem[Damen et~al.(2022)Damen, Doughty, Farinella, Furnari, Kazakos, Ma, Moltisanti, Munro, Perrett, Price, et~al.]{damen2022rescaling}
Dima Damen, Hazel Doughty, Giovanni~Maria Farinella, Antonino Furnari, Evangelos Kazakos, Jian Ma, Davide Moltisanti, Jonathan Munro, Toby Perrett, Will Price, et~al.
\newblock Rescaling egocentric vision: Collection, pipeline and challenges for epic-kitchens-100.
\newblock \emph{International Journal of Computer Vision}, pages 1--23, 2022.

\bibitem[Du et~al.(2024)Du, Yang, Dai, Dai, Nachum, Tenenbaum, Schuurmans, and Abbeel]{du2024learning}
Yilun Du, Sherry Yang, Bo Dai, Hanjun Dai, Ofir Nachum, Josh Tenenbaum, Dale Schuurmans, and Pieter Abbeel.
\newblock Learning universal policies via text-guided video generation.
\newblock \emph{Advances in Neural Information Processing Systems}, 36, 2024.

\bibitem[Esser et~al.(2023)Esser, Chiu, Atighehchian, Granskog, and Germanidis]{esser2023structure}
Patrick Esser, Johnathan Chiu, Parmida Atighehchian, Jonathan Granskog, and Anastasis Germanidis.
\newblock Structure and content-guided video synthesis with diffusion models.
\newblock In \emph{Proceedings of the IEEE/CVF International Conference on Computer Vision}, pages 7346--7356, 2023.

\bibitem[Fu et~al.(2025)Fu, Chen, Asad, Yuan, Oh, and Slabaugh]{fu2025handrawerleveragingspatialinformation}
Qifan Fu, Xu Chen, Muhammad Asad, Shanxin Yuan, Changjae Oh, and Gregory Slabaugh.
\newblock Handrawer: Leveraging spatial information to render realistic hands using a conditional diffusion model in single stage, 2025.

\bibitem[Grauman et~al.(2022)Grauman, Westbury, Byrne, Chavis, Furnari, Girdhar, Hamburger, Jiang, Liu, Liu, et~al.]{grauman2022ego4d}
Kristen Grauman, Andrew Westbury, Eugene Byrne, Zachary Chavis, Antonino Furnari, Rohit Girdhar, Jackson Hamburger, Hao Jiang, Miao Liu, Xingyu Liu, et~al.
\newblock Ego4d: Around the world in 3,000 hours of egocentric video.
\newblock In \emph{Proceedings of the IEEE/CVF Conference on Computer Vision and Pattern Recognition}, pages 18995--19012, 2022.

\bibitem[Grauman et~al.(2024)Grauman, Westbury, Torresani, Kitani, Malik, Afouras, Ashutosh, Baiyya, Bansal, Boote, et~al.]{grauman2024ego}
Kristen Grauman, Andrew Westbury, Lorenzo Torresani, Kris Kitani, Jitendra Malik, Triantafyllos Afouras, Kumar Ashutosh, Vijay Baiyya, Siddhant Bansal, Bikram Boote, et~al.
\newblock Ego-exo4d: Understanding skilled human activity from first-and third-person perspectives.
\newblock In \emph{Proceedings of the IEEE/CVF Conference on Computer Vision and Pattern Recognition}, pages 19383--19400, 2024.

\bibitem[Heusel et~al.(2017)Heusel, Ramsauer, Unterthiner, Nessler, and Hochreiter]{heusel2017gans}
Martin Heusel, Hubert Ramsauer, Thomas Unterthiner, Bernhard Nessler, and Sepp Hochreiter.
\newblock Gans trained by a two time-scale update rule converge to a local nash equilibrium.
\newblock \emph{Advances in neural information processing systems}, 30, 2017.

\bibitem[Ho et~al.(2020)Ho, Jain, and Abbeel]{ho2020denoising}
Jonathan Ho, Ajay Jain, and Pieter Abbeel.
\newblock Denoising diffusion probabilistic models.
\newblock \emph{Advances in neural information processing systems}, 33:\penalty0 6840--6851, 2020.

\bibitem[Ho et~al.(2022)Ho, Salimans, Gritsenko, Chan, Norouzi, and Fleet]{ho2022video}
Jonathan Ho, Tim Salimans, Alexey Gritsenko, William Chan, Mohammad Norouzi, and David~J Fleet.
\newblock Video diffusion models.
\newblock \emph{Advances in Neural Information Processing Systems}, 35:\penalty0 8633--8646, 2022.

\bibitem[Huang et~al.(2023)Huang, Sun, Xie, Li, and Liu]{huang2023t2i}
Kaiyi Huang, Kaiyue Sun, Enze Xie, Zhenguo Li, and Xihui Liu.
\newblock T2i-compbench: A comprehensive benchmark for open-world compositional text-to-image generation.
\newblock \emph{Advances in Neural Information Processing Systems}, 36:\penalty0 78723--78747, 2023.

\bibitem[Huang et~al.(2024{\natexlab{a}})Huang, He, Yu, Zhang, Si, Jiang, Zhang, Wu, Jin, Chanpaisit, et~al.]{huang2024vbench}
Ziqi Huang, Yinan He, Jiashuo Yu, Fan Zhang, Chenyang Si, Yuming Jiang, Yuanhan Zhang, Tianxing Wu, Qingyang Jin, Nattapol Chanpaisit, et~al.
\newblock Vbench: Comprehensive benchmark suite for video generative models.
\newblock In \emph{Proceedings of the IEEE/CVF Conference on Computer Vision and Pattern Recognition}, pages 21807--21818, 2024{\natexlab{a}}.

\bibitem[Huang et~al.(2024{\natexlab{b}})Huang, Zhang, Xu, He, Yu, Dong, Ma, Chanpaisit, Si, Jiang, et~al.]{huang2024vbench++}
Ziqi Huang, Fan Zhang, Xiaojie Xu, Yinan He, Jiashuo Yu, Ziyue Dong, Qianli Ma, Nattapol Chanpaisit, Chenyang Si, Yuming Jiang, et~al.
\newblock Vbench++: Comprehensive and versatile benchmark suite for video generative models.
\newblock \emph{arXiv preprint arXiv:2411.13503}, 2024{\natexlab{b}}.

\bibitem[Jain et~al.(2024)Jain, Attarian, Joshi, Wahid, Driess, Vuong, Sanketi, Sermanet, Welker, Chan, Gilitschenski, Bisk, and Dwibedi]{jain2024vid2robot}
Vidhi Jain, Maria Attarian, Nikhil~J Joshi, Ayzaan Wahid, Danny Driess, Quan Vuong, Pannag~R Sanketi, Pierre Sermanet, Stefan Welker, Christine Chan, Igor Gilitschenski, Yonatan Bisk, and Debidatta Dwibedi.
\newblock Vid2robot: End-to-end video-conditioned policy learning with cross-attention transformers, 2024.

\bibitem[Lai et~al.(2023)Lai, Dai, Chen, Pang, Rehg, and Liu]{lai2023lego}
Bolin Lai, Xiaoliang Dai, Lawrence Chen, Guan Pang, James~M Rehg, and Miao Liu.
\newblock Lego: Learning egocentric action frame generation via visual instruction tuning.
\newblock \emph{arXiv preprint arXiv:2312.03849}, 2023.

\bibitem[Lea et~al.(2017)Lea, Flynn, Vidal, Reiter, and Hager]{lea2017temporal}
Colin Lea, Michael~D Flynn, Rene Vidal, Austin Reiter, and Gregory~D Hager.
\newblock Temporal convolutional networks for action segmentation and detection.
\newblock In \emph{proceedings of the IEEE Conference on Computer Vision and Pattern Recognition}, pages 156--165, 2017.

\bibitem[LeCun et~al.(1998)LeCun, Bottou, Bengio, and Haffner]{lecun1998gradient}
Yann LeCun, L{\'e}on Bottou, Yoshua Bengio, and Patrick Haffner.
\newblock Gradient-based learning applied to document recognition.
\newblock \emph{Proceedings of the IEEE}, 86\penalty0 (11):\penalty0 2278--2324, 1998.

\bibitem[Lee et~al.(2024)Lee, Yasunaga, Meng, Mai, Park, Gupta, Zhang, Narayanan, Teufel, Bellagente, et~al.]{lee2024holistic}
Tony Lee, Michihiro Yasunaga, Chenlin Meng, Yifan Mai, Joon~Sung Park, Agrim Gupta, Yunzhi Zhang, Deepak Narayanan, Hannah Teufel, Marco Bellagente, et~al.
\newblock Holistic evaluation of text-to-image models.
\newblock \emph{Advances in Neural Information Processing Systems}, 36, 2024.

\bibitem[Li et~al.(2022)Li, Li, Xiong, and Hoi]{li2022blip}
Junnan Li, Dongxu Li, Caiming Xiong, and Steven Hoi.
\newblock Blip: Bootstrapping language-image pre-training for unified vision-language understanding and generation.
\newblock In \emph{International conference on machine learning}, pages 12888--12900. PMLR, 2022.

\bibitem[Lin et~al.(2022)Lin, Wang, Soldan, Wray, Yan, Xu, Gao, Tu, Zhao, Kong, et~al.]{lin2022egocentric}
Kevin~Qinghong Lin, Jinpeng Wang, Mattia Soldan, Michael Wray, Rui Yan, Eric~Z Xu, Difei Gao, Rong-Cheng Tu, Wenzhe Zhao, Weijie Kong, et~al.
\newblock Egocentric video-language pretraining.
\newblock \emph{Advances in Neural Information Processing Systems}, 35:\penalty0 7575--7586, 2022.

\bibitem[Lu et~al.(2022{\natexlab{a}})Lu, Zhou, Bao, Chen, Li, and Zhu]{lu2022dpm}
Cheng Lu, Yuhao Zhou, Fan Bao, Jianfei Chen, Chongxuan Li, and Jun Zhu.
\newblock Dpm-solver: A fast ode solver for diffusion probabilistic model sampling in around 10 steps.
\newblock \emph{Advances in Neural Information Processing Systems}, 35:\penalty0 5775--5787, 2022{\natexlab{a}}.

\bibitem[Lu et~al.(2022{\natexlab{b}})Lu, Zhou, Bao, Chen, Li, and Zhu]{lu2022dpmpp}
Cheng Lu, Yuhao Zhou, Fan Bao, Jianfei Chen, Chongxuan Li, and Jun Zhu.
\newblock Dpm-solver++: Fast solver for guided sampling of diffusion probabilistic models.
\newblock \emph{arXiv preprint arXiv:2211.01095}, 2022{\natexlab{b}}.

\bibitem[Lu et~al.(2024)Lu, Xu, Zhang, Wang, and Tao]{lu2024handrefiner}
Wenquan Lu, Yufei Xu, Jing Zhang, Chaoyue Wang, and Dacheng Tao.
\newblock Handrefiner: Refining malformed hands in generated images by diffusion-based conditional inpainting.
\newblock In \emph{ACM Multimedia 2024}, 2024.

\bibitem[Lugaresi et~al.(2019)Lugaresi, Tang, Nash, McClanahan, Uboweja, Hays, Zhang, Chang, Yong, Lee, et~al.]{lugaresi2019mediapipe}
Camillo Lugaresi, Jiuqiang Tang, Hadon Nash, Chris McClanahan, Esha Uboweja, Michael Hays, Fan Zhang, Chuo-Ling Chang, Ming~Guang Yong, Juhyun Lee, et~al.
\newblock Mediapipe: A framework for building perception pipelines.
\newblock \emph{arXiv preprint arXiv:1906.08172}, 2019.

\bibitem[Mandikal and Grauman(2021)]{mandikal2021dexvip}
Priyanka Mandikal and Kristen Grauman.
\newblock Dexvip: Learning dexterous grasping with human hand pose priors from video.
\newblock In \emph{Conference on Robot Learning}, 2021.

\bibitem[Marin et~al.(2018)Marin, Biswas, Ofli, Hynes, Salvador, Aytar, Weber, and Torralba]{marin2018recipe1m+}
Javier Marin, Aritro Biswas, Ferda Ofli, Nicholas Hynes, Amaia Salvador, Yusuf Aytar, Ingmar Weber, and Antonio Torralba.
\newblock Recipe1m+: a dataset for learning cross-modal embeddings for cooking recipes and food images.
\newblock \emph{arXiv preprint arXiv:1810.06553}, 2018.

\bibitem[Miech et~al.(2019)Miech, Zhukov, Alayrac, Tapaswi, Laptev, and Sivic]{miech2019howto100m}
Antoine Miech, Dimitri Zhukov, Jean-Baptiste Alayrac, Makarand Tapaswi, Ivan Laptev, and Josef Sivic.
\newblock Howto100m: Learning a text-video embedding by watching hundred million narrated video clips.
\newblock In \emph{Proceedings of the IEEE/CVF international conference on computer vision}, pages 2630--2640, 2019.

\bibitem[Narasimhaswamy et~al.(2024)Narasimhaswamy, Bhattacharya, Chen, Dasgupta, Mitra, and Hoai]{narasimhaswamy2024handiffuser}
Supreeth Narasimhaswamy, Uttaran Bhattacharya, Xiang Chen, Ishita Dasgupta, Saayan Mitra, and Minh Hoai.
\newblock Handiffuser: Text-to-image generation with realistic hand appearances.
\newblock In \emph{Proceedings of the IEEE/CVF Conference on Computer Vision and Pattern Recognition}, pages 2468--2479, 2024.

\bibitem[Ni et~al.(2023)Ni, Shi, Li, Huang, and Min]{ni2023conditional}
Haomiao Ni, Changhao Shi, Kai Li, Sharon~X Huang, and Martin~Renqiang Min.
\newblock Conditional image-to-video generation with latent flow diffusion models.
\newblock In \emph{Proceedings of the IEEE/CVF Conference on Computer Vision and Pattern Recognition}, pages 18444--18455, 2023.

\bibitem[Ni et~al.(2024)Ni, Egger, Lohit, Cherian, Wang, Koike-Akino, Huang, and Marks]{ni2024ti2v}
Haomiao Ni, Bernhard Egger, Suhas Lohit, Anoop Cherian, Ye Wang, Toshiaki Koike-Akino, Sharon~X Huang, and Tim~K Marks.
\newblock Ti2v-zero: Zero-shot image conditioning for text-to-video diffusion models.
\newblock In \emph{Proceedings of the IEEE/CVF Conference on Computer Vision and Pattern Recognition}, pages 9015--9025, 2024.

\bibitem[Nichol and Dhariwal(2021)]{nichol2021improved}
Alexander~Quinn Nichol and Prafulla Dhariwal.
\newblock Improved denoising diffusion probabilistic models.
\newblock In \emph{International conference on machine learning}, pages 8162--8171. PMLR, 2021.

\bibitem[Radford et~al.(2021)Radford, Kim, Hallacy, Ramesh, Goh, Agarwal, Sastry, Askell, Mishkin, Clark, et~al.]{radford2021learning}
Alec Radford, Jong~Wook Kim, Chris Hallacy, Aditya Ramesh, Gabriel Goh, Sandhini Agarwal, Girish Sastry, Amanda Askell, Pamela Mishkin, Jack Clark, et~al.
\newblock Learning transferable visual models from natural language supervision.
\newblock In \emph{International conference on machine learning}, pages 8748--8763. PMLR, 2021.

\bibitem[Ragusa et~al.(2021)Ragusa, Furnari, Livatino, and Farinella]{ragusa2021meccano}
Francesco Ragusa, Antonino Furnari, Salvatore Livatino, and Giovanni~Maria Farinella.
\newblock The meccano dataset: Understanding human-object interactions from egocentric videos in an industrial-like domain.
\newblock In \emph{Proceedings of the IEEE/CVF Winter Conference on Applications of Computer Vision}, pages 1569--1578, 2021.

\bibitem[Rombach et~al.(2022)Rombach, Blattmann, Lorenz, Esser, and Ommer]{rombach2022high}
Robin Rombach, Andreas Blattmann, Dominik Lorenz, Patrick Esser, and Bj{\"o}rn Ommer.
\newblock High-resolution image synthesis with latent diffusion models.
\newblock In \emph{Proceedings of the IEEE/CVF conference on computer vision and pattern recognition}, pages 10684--10695, 2022.

\bibitem[Saharia et~al.(2022)Saharia, Chan, Saxena, Li, Whang, Denton, Ghasemipour, Gontijo~Lopes, Karagol~Ayan, Salimans, et~al.]{saharia2022photorealistic}
Chitwan Saharia, William Chan, Saurabh Saxena, Lala Li, Jay Whang, Emily~L Denton, Kamyar Ghasemipour, Raphael Gontijo~Lopes, Burcu Karagol~Ayan, Tim Salimans, et~al.
\newblock Photorealistic text-to-image diffusion models with deep language understanding.
\newblock \emph{Advances in neural information processing systems}, 35:\penalty0 36479--36494, 2022.

\bibitem[Schuhmann et~al.(2021)Schuhmann, Vencu, Beaumont, Kaczmarczyk, Mullis, Katta, Coombes, Jitsev, and Komatsuzaki]{schuhmann2021laion}
Christoph Schuhmann, Richard Vencu, Romain Beaumont, Robert Kaczmarczyk, Clayton Mullis, Aarush Katta, Theo Coombes, Jenia Jitsev, and Aran Komatsuzaki.
\newblock Laion-400m: Open dataset of clip-filtered 400 million image-text pairs.
\newblock \emph{arXiv preprint arXiv:2111.02114}, 2021.

\bibitem[Sener et~al.(2022)Sener, Chatterjee, Shelepov, He, Singhania, Wang, and Yao]{sener2022assembly101}
Fadime Sener, Dibyadip Chatterjee, Daniel Shelepov, Kun He, Dipika Singhania, Robert Wang, and Angela Yao.
\newblock Assembly101: A large-scale multi-view video dataset for understanding procedural activities.
\newblock In \emph{Proceedings of the IEEE/CVF Conference on Computer Vision and Pattern Recognition}, pages 21096--21106, 2022.

\bibitem[Singer et~al.(2022)Singer, Polyak, Hayes, Yin, An, Zhang, Hu, Yang, Ashual, Gafni, et~al.]{singer2022make}
Uriel Singer, Adam Polyak, Thomas Hayes, Xi Yin, Jie An, Songyang Zhang, Qiyuan Hu, Harry Yang, Oron Ashual, Oran Gafni, et~al.
\newblock Make-a-video: Text-to-video generation without text-video data.
\newblock \emph{arXiv preprint arXiv:2209.14792}, 2022.

\bibitem[Smith et~al.(2018)Smith, Toor, and Van~Kessel]{smith2018many}
Aaron Smith, Skye Toor, and Patrick Van~Kessel.
\newblock Many turn to youtube for children’s content, news, how-to lessons.
\newblock \emph{Pew Research Center}, 7, 2018.

\bibitem[Song et~al.(2020)Song, Meng, and Ermon]{song2020denoising}
Jiaming Song, Chenlin Meng, and Stefano Ermon.
\newblock Denoising diffusion implicit models.
\newblock \emph{arXiv preprint arXiv:2010.02502}, 2020.

\bibitem[Soni et~al.(2024)Soni, Venkataraman, Chandra, Fischmeister, Liang, Dai, and Yang]{soni2024videoagent}
Achint Soni, Sreyas Venkataraman, Abhranil Chandra, Sebastian Fischmeister, Percy Liang, Bo Dai, and Sherry Yang.
\newblock Videoagent: Self-improving video generation.
\newblock \emph{arXiv preprint arXiv:2410.10076}, 2024.

\bibitem[Sou{\v{c}}ek et~al.(2024)Sou{\v{c}}ek, Damen, Wray, Laptev, and Sivic]{souvcek2024genhowto}
Tom{\'a}{\v{s}} Sou{\v{c}}ek, Dima Damen, Michael Wray, Ivan Laptev, and Josef Sivic.
\newblock Genhowto: Learning to generate actions and state transformations from instructional videos.
\newblock In \emph{2024 IEEE/CVF Conference on Computer Vision and Pattern Recognition (CVPR)}, pages 6561--6571. IEEE, 2024.

\bibitem[Unterthiner et~al.(2019)Unterthiner, van Steenkiste, Kurach, Marinier, Michalski, and Gelly]{unterthiner2019fvd}
Thomas Unterthiner, Sjoerd van Steenkiste, Karol Kurach, Raphaël Marinier, Marcin Michalski, and Sylvain Gelly.
\newblock Fvd: A new metric for video generation.
\newblock In \emph{ICLR Workshop on Deep Generative Models for Structured Data (DeepGenStruct)}, 2019.

\bibitem[Varol et~al.(2017)Varol, Laptev, and Schmid]{varol2017long}
G{\"u}l Varol, Ivan Laptev, and Cordelia Schmid.
\newblock Long-term temporal convolutions for action recognition.
\newblock \emph{IEEE transactions on pattern analysis and machine intelligence}, 40\penalty0 (6):\penalty0 1510--1517, 2017.

\bibitem[Vaswani(2017)]{vaswani2017attention}
A Vaswani.
\newblock Attention is all you need.
\newblock \emph{Advances in Neural Information Processing Systems}, 2017.

\bibitem[Wang et~al.(2023{\natexlab{a}})Wang, Saharia, Montgomery, Pont-Tuset, Noy, Pellegrini, Onoe, Laszlo, Fleet, Soricut, et~al.]{wang2023imagen}
Su Wang, Chitwan Saharia, Ceslee Montgomery, Jordi Pont-Tuset, Shai Noy, Stefano Pellegrini, Yasumasa Onoe, Sarah Laszlo, David~J Fleet, Radu Soricut, et~al.
\newblock Imagen editor and editbench: Advancing and evaluating text-guided image inpainting.
\newblock In \emph{Proceedings of the IEEE/CVF conference on computer vision and pattern recognition}, pages 18359--18369, 2023{\natexlab{a}}.

\bibitem[Wang et~al.()Wang, Kwon, Pan, Chakraborty, Andrist, Feniello, Tekin, Frujeri, and Pollefeys]{wangholoassist}
Xin Wang, Taein Kwon, Mahdi Rad1~Bowen Pan, Ishani Chakraborty, Sean Andrist, Dan Bohus1~Ashley Feniello, Bugra Tekin, Felipe~Vieira Frujeri, and Neel Joshi1~Marc Pollefeys.
\newblock Holoassist: an egocentric human interaction dataset for interactive ai assistants in the real world supplementary material.

\bibitem[Wang et~al.(2016)Wang, Ong, and Nee]{wang2016multi}
Xuan Wang, SK Ong, and Andrew Yeh-Ching Nee.
\newblock Multi-modal augmented-reality assembly guidance based on bare-hand interface.
\newblock \emph{Advanced Engineering Informatics}, 30\penalty0 (3):\penalty0 406--421, 2016.

\bibitem[Wang et~al.(2023{\natexlab{b}})Wang, Kwon, Rad, Pan, Chakraborty, Andrist, Bohus, Feniello, Tekin, Frujeri, et~al.]{wang2023holoassist}
Xin Wang, Taein Kwon, Mahdi Rad, Bowen Pan, Ishani Chakraborty, Sean Andrist, Dan Bohus, Ashley Feniello, Bugra Tekin, Felipe~Vieira Frujeri, et~al.
\newblock Holoassist: an egocentric human interaction dataset for interactive ai assistants in the real world.
\newblock In \emph{Proceedings of the IEEE/CVF International Conference on Computer Vision}, pages 20270--20281, 2023{\natexlab{b}}.

\bibitem[Wang et~al.(2024{\natexlab{a}})Wang, Yuan, Zhang, Chen, Wang, Zhang, Shen, Zhao, and Zhou]{wang2024videocomposer}
Xiang Wang, Hangjie Yuan, Shiwei Zhang, Dayou Chen, Jiuniu Wang, Yingya Zhang, Yujun Shen, Deli Zhao, and Jingren Zhou.
\newblock Videocomposer: Compositional video synthesis with motion controllability.
\newblock \emph{Advances in Neural Information Processing Systems}, 36, 2024{\natexlab{a}}.

\bibitem[Wang et~al.(2024{\natexlab{b}})Wang, Zhao, Liu, Wang, Zhao, Bao, Zhu, Zhang, and Wang]{wang2024egovid}
Xiaofeng Wang, Kang Zhao, Feng Liu, Jiayu Wang, Guosheng Zhao, Xiaoyi Bao, Zheng Zhu, Yingya Zhang, and Xingang Wang.
\newblock Egovid-5m: A large-scale video-action dataset for egocentric video generation.
\newblock \emph{arXiv preprint arXiv:2411.08380}, 2024{\natexlab{b}}.

\bibitem[Wang et~al.(2024{\natexlab{c}})Wang, Chen, Ma, Zhou, Huang, Wang, Yang, He, Yu, Yang, et~al.]{wang2024lavie}
Yaohui Wang, Xinyuan Chen, Xin Ma, Shangchen Zhou, Ziqi Huang, Yi Wang, Ceyuan Yang, Yinan He, Jiashuo Yu, Peiqing Yang, et~al.
\newblock Lavie: High-quality video generation with cascaded latent diffusion models.
\newblock \emph{International Journal of Computer Vision}, pages 1--20, 2024{\natexlab{c}}.

\bibitem[Wu et~al.(2024{\natexlab{a}})Wu, Fang, Wu, Wang, Ge, Cun, Zhang, Liu, Gu, Zhao, et~al.]{wu2024towards}
Jay~Zhangjie Wu, Guian Fang, Haoning Wu, Xintao Wang, Yixiao Ge, Xiaodong Cun, David~Junhao Zhang, Jia-Wei Liu, Yuchao Gu, Rui Zhao, et~al.
\newblock Towards a better metric for text-to-video generation.
\newblock \emph{arXiv preprint arXiv:2401.07781}, 2024{\natexlab{a}}.

\bibitem[Wu et~al.(2024{\natexlab{b}})Wu, Si, Jiang, Huang, and Liu]{wu2024freeinit}
Tianxing Wu, Chenyang Si, Yuming Jiang, Ziqi Huang, and Ziwei Liu.
\newblock Freeinit: Bridging initialization gap in video diffusion models.
\newblock In \emph{European Conference on Computer Vision}, pages 378--394, 2024{\natexlab{b}}.

\bibitem[Xing et~al.(2023)Xing, Xia, Zhang, Chen, Wang, Wong, and Shan]{xing2023dynamicrafter}
Jinbo Xing, Menghan Xia, Yong Zhang, Haoxin Chen, Xintao Wang, Tien-Tsin Wong, and Ying Shan.
\newblock Dynamicrafter: Animating open-domain images with video diffusion priors.
\newblock 2023.

\bibitem[Xu et~al.(2025)Xu, Huang, Pei, Hou, Li, Chen, Zhang, Feng, and Xie]{jilanxgen2025}
Jilan Xu, Yifei Huang, Baoqi Pei, Junlin Hou, Qingqiu Li, Guo Chen, Yuejie Zhang, Rui Feng, and Weidi Xie.
\newblock X-gen: Ego-centric video prediction by watching exo-centric videos.
\newblock In \emph{The Thirteenth International Conference on Learning Representations}, 2025.

\bibitem[Xue et~al.(2022)Xue, Hang, Zeng, Sun, Liu, Yang, Fu, and Guo]{xue2022advancing}
Hongwei Xue, Tiankai Hang, Yanhong Zeng, Yuchong Sun, Bei Liu, Huan Yang, Jianlong Fu, and Baining Guo.
\newblock Advancing high-resolution video-language representation with large-scale video transcriptions.
\newblock In \emph{Proceedings of the IEEE/CVF Conference on Computer Vision and Pattern Recognition}, pages 5036--5045, 2022.

\bibitem[Yan et~al.(2019)Yan, Tu, Wang, Zhang, Hao, Zhang, and Dai]{yan2019stat}
Chenggang Yan, Yunbin Tu, Xingzheng Wang, Yongbing Zhang, Xinhong Hao, Yongdong Zhang, and Qionghai Dai.
\newblock Stat: Spatial-temporal attention mechanism for video captioning.
\newblock \emph{IEEE transactions on multimedia}, 22\penalty0 (1):\penalty0 229--241, 2019.

\bibitem[Yang et~al.(2023)Yang, Du, Ghasemipour, Tompson, Schuurmans, and Abbeel]{yang2023learning}
Mengjiao Yang, Yilun Du, Kamyar Ghasemipour, Jonathan Tompson, Dale Schuurmans, and Pieter Abbeel.
\newblock Learning interactive real-world simulators.
\newblock \emph{arXiv preprint arXiv:2310.06114}, 2023.

\bibitem[Yang et~al.(2024{\natexlab{a}})Yang, Gandhi, and Turk]{yang2024annotated}
Yue Yang, Atith~N Gandhi, and Greg Turk.
\newblock Annotated hands for generative models.
\newblock \emph{arXiv preprint arXiv:2401.15075}, 2024{\natexlab{a}}.

\bibitem[Yang et~al.(2024{\natexlab{b}})Yang, Teng, Zheng, Ding, Huang, Xu, Yang, Hong, Zhang, Feng, et~al.]{yang2024cogvideox}
Zhuoyi Yang, Jiayan Teng, Wendi Zheng, Ming Ding, Shiyu Huang, Jiazheng Xu, Yuanming Yang, Wenyi Hong, Xiaohan Zhang, Guanyu Feng, et~al.
\newblock Cogvideox: Text-to-video diffusion models with an expert transformer.
\newblock \emph{arXiv preprint arXiv:2408.06072}, 2024{\natexlab{b}}.

\bibitem[Yilun et~al.(2023)Yilun, Mengjiao, Bo, Hanjun, Ofir, B, Schuurmans, and Abbeel]{du2023learning}
Du Yilun, Yang Mengjiao, Dai Bo, Dai Hanjun, Nachum Ofir, Tenenbaum~Joshua B, Dale Schuurmans, and Pieter Abbeel.
\newblock Learning universal policies via text-guided video generation.
\newblock \emph{arXiv e-prints}, pages arXiv--2302, 2023.

\bibitem[Yin et~al.(2023)Yin, Wu, Liang, Shi, Li, Ming, and Duan]{yin2023dragnuwa}
Shengming Yin, Chenfei Wu, Jian Liang, Jie Shi, Houqiang Li, Gong Ming, and Nan Duan.
\newblock Dragnuwa: Fine-grained control in video generation by integrating text, image, and trajectory.
\newblock \emph{arXiv preprint arXiv:2308.08089}, 2023.

\bibitem[Zhang et~al.(2024{\natexlab{a}})Zhang, Wu, Liu, Zhao, Ran, Gu, Gao, and Shou]{zhang2024show1}
D.~J. Zhang, J.~Z. Wu, J.-W. Liu, R. Zhao, L. Ran, Y. Gu, D. Gao, and M.~Z. Shou.
\newblock Show-1: Marrying pixel and latent diffusion models for text-to-video generation.
\newblock \emph{International Journal of Computer Vision}, 2024{\natexlab{a}}.

\bibitem[Zhang et~al.(2022)Zhang, Cai, Pan, Hong, Guo, Yang, and Liu]{zhang2022motiondiffuse}
Mingyuan Zhang, Zhongang Cai, Liang Pan, Fangzhou Hong, Xinying Guo, Lei Yang, and Ziwei Liu.
\newblock Motiondiffuse: Text-driven human motion generation with diffusion model.
\newblock \emph{arXiv preprint arXiv:2208.15001}, 2022.

\bibitem[Zhang et~al.(2024{\natexlab{b}})Zhang, Xing, Zeng, Fang, and Chen]{zhang2024pia}
Yiming Zhang, Zhening Xing, Yanhong Zeng, Youqing Fang, and Kai Chen.
\newblock Pia: Your personalized image animator via plug-and-play modules in text-to-image models.
\newblock In \emph{Proceedings of the IEEE/CVF Conference on Computer Vision and Pattern Recognition}, pages 7747--7756, 2024{\natexlab{b}}.

\bibitem[Zheng et~al.(2024)Zheng, Peng, Yang, Shen, Li, Liu, Zhou, Li, and You]{opensora}
Zangwei Zheng, Xiangyu Peng, Tianji Yang, Chenhui Shen, Shenggui Li, Hongxin Liu, Yukun Zhou, Tianyi Li, and Yang You.
\newblock Open-sora: Democratizing efficient video production for all, 2024.

\bibitem[Zhou et~al.(2018)Zhou, Xu, and Corso]{zhou2018towards}
Luowei Zhou, Chenliang Xu, and Jason Corso.
\newblock Towards automatic learning of procedures from web instructional videos.
\newblock In \emph{Proceedings of the AAAI Conference on Artificial Intelligence}, 2018.

\bibitem[Zhou et~al.(2024)Zhou, Zhang, Yang, Qian, and Li]{zhou2024motion}
Qiang Zhou, Shaofeng Zhang, Nianzu Yang, Ye Qian, and Hao Li.
\newblock Motion control for enhanced complex action video generation.
\newblock \emph{arXiv preprint arXiv:2411.08328}, 2024.

\end{thebibliography}
}

\newpage
\appendix

\section{BLIP Score with Different BLIP Model Size}
BLIP Score~\cite{lai2023lego} leverage pre-trained BLIP model~\cite{li2022blip} to evaluate how close the generated video content is to the target action text description. In the main script, we introduce the evaluation results using the large BLIP model. In this supplementary material, we provide evaluation results from the small BLIP model as well. These results yield the same conclusion as we have in the main script --- our method outperforms other baseline methods.

\begin{table}[h!]
    \centering
    \small
    \begin{tabular}{
    cccccc}
        \toprule
        \multirow{1}{*}{\textbf{Dataset}} & 
        \multirow{1}{*}{\textbf{Method}} & 
        \textbf{BLIP\_large $\uparrow$} & 
        \textbf{BLIP\_base $\uparrow$} 
        \\
        
        \midrule
        \multirow{5}{*}{\textbf{EpicKitchens}} 
        & LFDM~\cite{ni2023conditional} & 0.235 & 0.223\\
        & AA~\cite{dai2023fine} & 0.295 & 0.298\\
        & AVDC~\cite{du2023learning} & 0.116 & 0.106\\
        & PIA~\cite{zhang2024pia} &  0.294 & 0.299\\
        & HANDI & \textbf{0.298} & \textbf{0.335}\\
        \midrule
        \multirow{5}{*}{\textbf{Ego4D}} 
        & LFDM~\cite{ni2023conditional} & 0.221 & 0.219\\
        & AA~\cite{dai2023fine} & 0.260 & 0.264\\
        & AVDC~\cite{du2023learning} & 0.155 & 0.161\\
        & PIA~\cite{zhang2024pia} & 0.219 & 0.260\\
        & HANDI & \textbf{0.263} & \textbf{0.306}\\
        \bottomrule
    \end{tabular}
    \caption{Quantitative results on EpicKitchens~\cite{damen2022rescaling}, Ego4D~\cite{grauman2022ego4d} and a Motion Intensive subset of EpicKitchens. Our method outperforms all baselines across all metrics on at least one dataset.}
    \label{tab:main }
\end{table}

\section{Generation of Various Actions in the Same Visual Context}
We explore our model's performance on generating different actions given the same input context image. These are novel image-action pairs that have never shown in the dataset including the test set. We illustrate one example in~\cref{fig:various_prompt} that crosses two existing samples in the test set. 
These results further indicate that our model can generate Hand-Centric videos for various actions, emphasizing hand movement and subtle fingertip motion.

\begin{TODOB}
\section{Training on Camera-Movement-Reduced Dataset}
This section needs to be wrote up with method description on reducing cam-mov and the experiment results (quantitative and qualitative)

In supplementary, add the best qualitative results from the Cam-Red trained model.

\section{Foundational Discussion on Diffusion Models}
\paragraph{DDPM~\cite{ho2020denoising} vs. DDIM~\cite{song2020denoising}: towards how exactly the noise adding process work.}  Denoising Diffusion Implicit Models (DDIM) provide a deterministic alternative to the stochastic sampling process in Denoising Diffusion Probabilistic Models (DDPM), enabling significantly faster generation without substantial degradation in sample quality. In DDPM, the generative process is a Markovian chain with Gaussian transitions, where at each reverse step, noise is sampled from a learned distribution (the trained model outputs the mean of the distribution):
\begin{align}
p_\theta(x_{t-1} \mid x_t) = \mathcal{N}(x_{t-1}; \mu_\theta(x_t, t), \sigma_t^2 I),
\end{align}
making the sampling process inherently stochastic despite the model predicting the mean via \( \epsilon_\theta(x_t, t) \). In contrast, DDIM reparameterizes the reverse dynamics to eliminate this stochasticity, yielding a non-Markovian and often deterministic process. Instead of sampling, DDIM uses:
\begin{align}
x_{t-1} = \sqrt{\alpha_{t-1}} \left( \frac{x_t - \sqrt{1 - \alpha_t} \cdot \epsilon_\theta}{\sqrt{\alpha_t}} \right) + \eta \cdot \sigma_t \cdot z,
\end{align}
where \( \eta \in [0, 1] \) controls the level of added noise, and \( \eta = 0 \) results in a fully deterministic trajectory. This enables sampling in far fewer steps (e.g., 10–50 vs. 1000 in DDPM), facilitating applications such as fast generation, latent interpolation, and image editing, albeit with slightly reduced output diversity.

\section{Denosing Process Visualization} In supplementary, show the denoising process of our model in video.

\section{Discussion on gap between training and inference insolver.~\cite{huang2024vbench,huang2024vbench++}}

\section{Discussion on initialization of the diffusion model.~\cite{wu2024freeinit}}

\section{Post-Publishing Related Work}
EgoVid-5M is an instructional video dataset tailored for egocentric video generation~\cite{wang2024egovid}. They provide fine-grained annotation like kinemetic control, discussed camera movement and etc.. 

MVideo is not an TI2V work but discuss a method for improving Complex Action Video Generation~\cite{zhou2024motion}. 

X-Gen~\cite{jilanxgen2025} is a work generating Ego-Centric Video --- the same output as HANDI --- from Exo-Centric Video (and text?) --- different input information. From problem setting standpoint, HCVG is a more challenging problem. Several aspects we can learn from them are: 
\begin{itemize}
    \item We struggled with reasoning specialty in ego-centric videos, among other normal video generation work. That is why we in the end put the angle as ``hand-centric''. X-Gen puts the viewpoint directly as the angle.
    \item Technically, this problem is likely rendered as normal conditional video generation work. We had the same challenge of putting technical contribution beyond problem setting contribution. How does X-Gen address the technical novelty? 
\end{itemize} 
\end{TODOB}

\end{document}